\documentclass{article}

\usepackage[main, final]{local_style}

\usepackage[utf8]{inputenc}
\usepackage[T1]{fontenc}
\usepackage{titletoc}
\usepackage{xcolor}
\usepackage{colortbl}
\usepackage{nicematrix}
\definecolor{paperteal}{HTML}{006a98}
\definecolor{sailblue}{HTML}{EBF4F8}
\newcolumntype{s}{>{\columncolor{sailblue}}c}
\usepackage{hyperref}
\hypersetup{
  colorlinks=true,
  linkcolor=paperteal,
  citecolor=paperteal,
  urlcolor=paperteal
}
\usepackage{url}
\usepackage{booktabs}
\usepackage{amsfonts}
\usepackage{nicefrac}
\usepackage{microtype}

\usepackage{array}
\usepackage{float}
\usepackage{graphicx}
\usepackage{amsmath}
\usepackage{wrapfig}
\usepackage{multirow}

\title{Transferable SCF-Acceleration through Solver-Aligned Initialization Learning}

\author{%
  Eike S.~Eberhard\thanks{Equal contribution}$^{\hspace{2mm} 1,2,3}$, \:
  Viktor Kotsev\footnotemark[1]$^{\hspace{2mm} 1,2}$, \:
  Timm Güthle$^{1}$, \:
  Stephan Günnemann$^{1,2,3}$\\[0.5em]
  \href{mailto:e.eberhard@tum.de,viktor.kotsev@tum.de}{\{e.eberhard, viktor.kotsev\}@tum.de}\\[0.3em]
  $^1$Technical University of Munich (TUM) \quad
  $^2$Munich Data Science Institute (MDSI) \\
  $^3$Munich Center for Machine Learning (MCML)
}

\begin{document}

\maketitle

\newcommand{\BxB}{\mathbb{R}^{B \times B}}
\newcommand{\Rthree}{\mathbb{R}^3}
\newcommand{\Gr}{\mathrm{Gr}}
\newcommand{\tr}{\mathrm{tr}}
\newcommand{\diag}{\mathrm{diag}}
\newcommand{\occ}{\mathrm{occ}}
\newcommand{\vir}{\mathrm{virt}}

\newcommand{\oursLong}{Solver-Aligned Initialization Learning}
\newcommand{\ours}{SAIL}

\begin{abstract}
    The cost of Kohn-Sham density functional theory (KS-DFT) calculations scales with the number of required solver iterations, which depend on the quality of the initial guess.
    Machine learning methods that predict initial guesses from molecular geometry can reduce this cost, but matrix-prediction models fail when extrapolating to larger molecules, degrading rather than accelerating convergence~\citep{liuUniversallyTransferableAcceleration2025}.
    We show that this failure is a supervision problem, not an extrapolation problem: models trained on ground-state targets fit those targets well out of distribution, yet produce initial guesses that slow convergence.
    \oursLong{} (\ours{}) resolves this for both Hamiltonian and density matrix models by differentiating through parts of the self-consistent field (SCF) solver end-to-end.
    We introduce the Effective Relative Iteration Count (ERIC), a correction to the commonly used RIC that accounts for hidden Fock-build overhead.
    On QM40, which contains molecules up to $4\times$ larger than the training distribution, \ours{} reduces ERIC by 40\% (PBE), 35\% (SCAN), and 32\% (B3LYP), more than doubling the previous state-of-the-art reduction on B3LYP.
    On QMugs molecules $10\times$ larger than the training set, \ours{} delivers a $1.35\times$ wall-time speedup at the hybrid level of theory, extending ML SCF acceleration to large drug-like molecules.
    A reference implementation is available at: \url{https://github.com/ESEberhard/ML4DFT}.
\end{abstract}

\section{Introduction}
    Self-consistent field (SCF) solvers are the computational backbone of Hartree-Fock (HF)~\citep{hartreeWaveMechanicsAtom1928,fockNaeherungsmethodeZurLoesung1930} and KS-DFT calculations~\citep{hohenbergInhomogeneousElectronGas1964,kohnSelfConsistentEquationsIncluding1965}. KS-DFT is the most widely used electronic structure method in computational chemistry~\citep{Szabo1996}, reflected in the hundreds of thousands of publications~\citep{haunschildComprehensiveAnalysisHistory2019} and the substantial HPC resources devoted to these calculations~\citep{antypasNERSCWorkloadAnalysis2014, zhaoVASPPerformanceHPE2023}.
    Any reliable method for speeding up SCF calculations increases resource efficiency and accelerates scientific discovery, both directly and through downstream applications such as machine-learned interatomic potentials (MLIPs), where the cost of generating large reference datasets remains a bottleneck~\citep{kulichenkoDataGenerationMachine2024a}. Moreover, unlike machine-learned surrogates that aim to replace DFT entirely, solver acceleration preserves accuracy, allowing practitioners to choose the level of theory \emph{and} its \textbf{known limitations} rather than adopt those of a yet-to-be-characterized surrogate.

    In recent years, ML initialization methods have been proposed to accelerate SCF calculations~\citep{kokerHigherorderEquivariantNeural2024, yuQH9QuantumHamiltonian2024, febrerGraph2MatUniversalGraph2025}.
    These methods produce initial guesses conditioned on the molecular point cloud and typically fall into one of three categories. We jointly refer to Hamiltonian and density matrix prediction as \textbf{matrix-based methods}, as both parameterize quadratic basis-set expansions with coefficient matrices $\mathbf{X} \in \mathbb{R}^{B \times B}$. The third category, the \textbf{coefficient-based} ansatz, predicts the coefficients $\mathbf{c}\in \mathbb{R}^{B_\textrm{aux}}$ of a linear auxiliary basis-set expansion.
    \citet{liuUniversallyTransferableAcceleration2025} have shown that matrix-based approaches \emph{decelerate} SCF calculations when \textbf{extrapolating to larger molecules}, a groundbreaking observation given that prior work on matrix-based methods suggested robust out-of-distribution (OOD) performance based on, in hindsight, insufficient extrapolation benchmarks. They therefore proposed adopting coefficient-based approaches~\citep{songNeuralSCFNeuralNetwork2024}, which are, by default, limited to local (GGA-type) functionals.
    Unfortunately, this level of theory is insufficient for many practical applications, particularly in organic chemistry, which is dominated by non-local (range-separated hybrid) functionals~\citep{isertQMugsQuantumMechanical2021, eastmanSPICEDatasetDruglike2023, madushankaQM40RealisticQuantum2024, levineOpenMolecules20252025}.

    In this work, we show that the currently prevailing approach of training initial-guess models by fitting ground-state quantities is suboptimal.
    We address this with \textbf{\oursLong{} (\ours{})}, which backpropagates through the SCF algorithm to the ML initial guess, training on solver dynamics rather than ground-state references. This procedure is label-free and requires only molecular geometries. We show that \ours{} cures the size-extrapolation issues for matrix-based methods and reveal a \textbf{misalignment between ground-state supervision and initial-guess quality}.
    For the extension of coefficient-based models to non-local functionals proposed by~\citet{liuUniversallyTransferableAcceleration2025}, the constant (non-learnable) treatment of the non-local Hamiltonian contribution makes ground-state supervision suboptimal by construction, \emph{even in distribution}. Training with \ours{} enables the model to learn a systematic density error compensating for this fixed approximation.

    \begin{figure}[t]
      \centering
      \includegraphics[width=\textwidth]{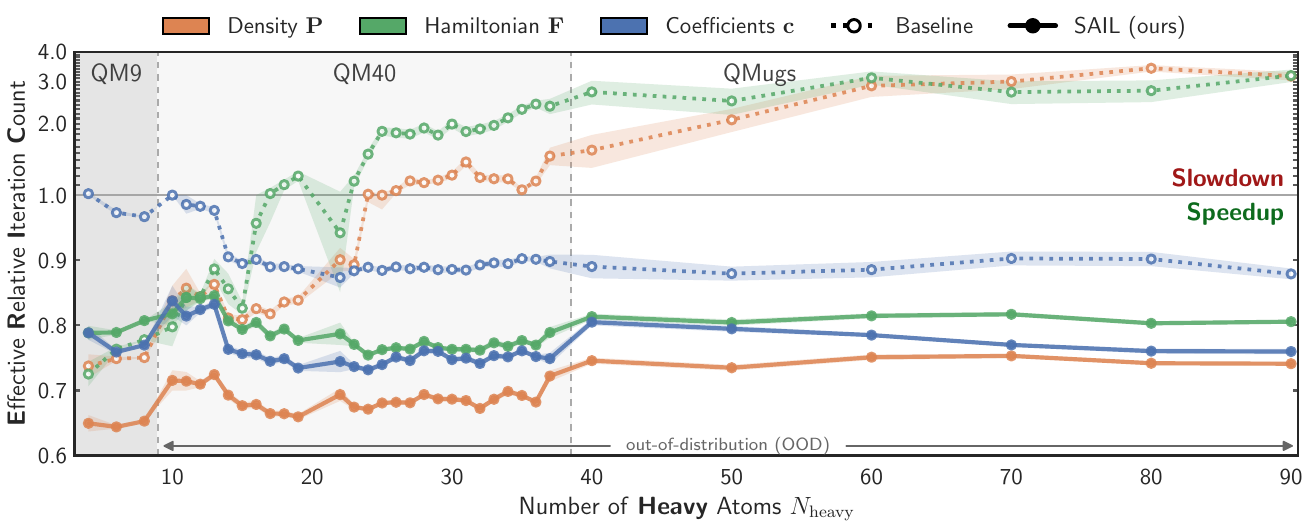}
      \vspace{-0.4cm}
      \caption{
            \textbf{Size extrapolation of \ours{} on B3LYP / def2-SVP.} ERIC of initial-guess models trained on QM9 and evaluated on QM9, QM40, and QMugs, with QMugs molecules up to $10\times$ larger than those seen during training. ERIC $< 1$ indicates speedup over the traditional initial guess, ERIC $> 1$ slowdown. Matrix-models trained on ground-state targets degrade out of distribution, whereas \ours{} maintains a $\sim 25\%$ speedup across the full range. The coefficient model is robust out of distribution, yet every \ours{}-trained model (matrix variants included) outperforms it across the full range.
        }
      \label{fig:Qmugs-B3LYP-eric}
    \end{figure}

\section{Background}
    KS-DFT describes the quantum mechanical ground state in terms of its electron density $\rho^* \in \mathcal{D}_{N_e} = \{\mathbb{R}^3 \rightarrow [0, \infty) | \int_{\mathbb{R}^3} \rho(\mathbf r) \:d\mathbf r = N_e\}$ and energy $E^*\in \mathbb{R}$ to determine most chemical properties of interest.
    Molecular DFT codes expand $\rho$ in a finite set $\{\chi_\mu\}_{\mu=1}^B$ of atom-centered basis functions
    \begin{align}
        \rho(\mathbf{r}) = \sum_{\mu\nu}^B \chi_\mu(\mathbf{r}) \, P_{\mu\nu} \, \chi_\nu(\mathbf{r})\:,
    \end{align}
    where $\mathbf P \in \BxB$ is the density matrix.
    In this representation, the ground-state energy functional $E[\rho]\colon \mathcal{D}_{N_e} \rightarrow \mathbb{R}$ reduces to a function $E(\mathbf{P})\colon \BxB \rightarrow \mathbb{R}$, and finding the ground state becomes an energy minimization problem within the space of valid density matrices~\citep{lehtolaOverviewSelfConsistentField2020}.
    SCF solvers approach this minimization problem iteratively. Each step computes the energy gradient $\mathbf{F} = \partial E / \partial \mathbf{P}$, the so-called Fock matrix, and obtains an updated density $\mathbf{P}^{(t+1)}$ by solving a generalized eigenvalue problem (Appendix~\ref{app:scf-details}):
    \begin{align}
        \label{eq:scf}
        \mathbf{P}^{(t+1)} = \mathrm{SCF}(\mathbf{P}^{(t)}) = \left(\mathcal{M}_{\mathbf{F}\rightarrow\mathbf{P}} \circ \mathcal{M}_{\mathbf{P}\rightarrow\mathbf{F}}\right)(\mathbf{P}^{(t)}) \:.
    \end{align}

    This map is applied until \emph{self-consistency} is reached, meaning $\mathbf{P}^{*} \simeq \mathrm{SCF}(\mathbf{P}^{*})$.
    In practice, convergence accelerators augment this fixed-point iteration and drastically alter its dynamics (Appendix~\ref{app:diis}), but the quality of the \textbf{initial guess} $\mathbf{P}^{(0)}$ remains critical. It determines \emph{both} whether the solver converges \emph{and} how many iterations it requires~\citep{lehtolaAssessmentInitialGuesses2019}. The standard initialization in most DFT codes is based on the superposition of atomic densities (SAD)~\citep{vanlentheStartingSCFCalculations2006}.

\begin{wrapfigure}[21]{r}{0.5\textwidth}
        \centering
        \vspace{-1mm}
        \includegraphics[width=\linewidth]{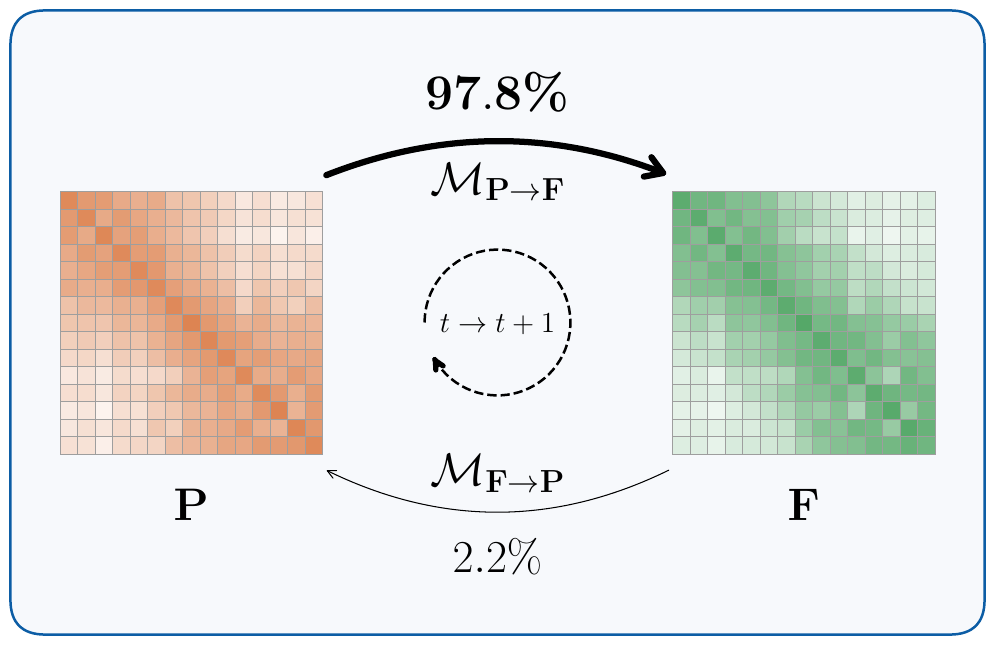}
        \vspace{-4mm}
        \caption{\textbf{SCF cycle wall-time analysis} at the B3LYP/def2-SVP level of theory, with jk-density-fitting. Each iteration constructs the Fock matrix $\mathbf{F}^{(t)} = \mathcal{M}_{\mathbf{P}\to\mathbf{F}}(\mathbf{P}^{(t)})$ and updates the density $\mathbf{P}^{(t+1)} = \mathcal{M}_{\mathbf{F}\to\mathbf{P}}(\mathbf{F}^{(t)})$, see Eq.~\eqref{eq:scf}. The Fock build dominates at 97.8\% of the iteration cost. Cost measured on a QM40 molecule with 30 heavy atoms using GPU4PySCF~\citep{liIntroducingGPUAcceleration2025}.}
        \label{fig:scf-cycle}
    \end{wrapfigure}

    Since wall-time reduction is the ultimate goal, understanding how the computational costs of an SCF step are distributed is essential.
    The \textbf{Fock build} $\mathcal{M}_{\mathbf{P}\rightarrow\mathbf{F}}$ dominates the cost (Fig.~\ref{fig:scf-cycle}).
    Its scaling depends on the choice of the exchange-correlation (XC) functional: $\mathcal{O}(B^3)$ for semi-local functionals such as PBE and SCAN, and $\mathcal{O}(B^4)$ for hybrid functionals like B3LYP~\citep{perdewJacobsLadderDensity2001}.

    \looseness=-1
    \textbf{Density fitting} lowers the scaling of the electrostatic contribution to the Fock build from $\mathcal{O}(B^4)$ to $\mathcal{O}(B^3)$ by approximating the electron density as $\rho(\mathbf{r}) \approx$  \scalebox{0.95}{$\sum_\mu^{B_\textrm{aux}} c_\mu \, \chi^{(\text{aux})}_\mu(\mathbf{r})$}, where $\mathbf{c} \in \mathbb{R}^{B_\text{aux}}$ are the fitting coefficients and $B_\text{aux} \ll B^2$~\citep{whittenCoulombicPotentialEnergy1973, vahtrasIntegralApproximationsLCAOSCF1993}.
    Apart from enabling the $\mathcal{O}(B^3)$ scaling of semi-local functionals, this expansion provides a natural interface between equivariant GNNs and the electron density. Since the auxiliary basis consists of equivariant functions centered on individual nuclei, the coefficients $\mathbf{c}$ are directly predictable from node-level representations~\citep{songNeuralSCFNeuralNetwork2024,liuUniversallyTransferableAcceleration2025}, unlike the density matrix representation, which couples basis functions across atom pairs.

\section{Related Work}
    \label{sec:related work}
    \textbf{Hamiltonian prediction} methods are trained to predict the converged Fock matrix\footnote{In finite-basis-set calculations, the Hamiltonian is commonly referred to as the Fock matrix} $\hat{\mathbf{F}} \approx \mathbf{F}^*$~\citep{schuttUnifyingMachineLearning2019,yuEfficientEquivariantGraph2023,yuQH9QuantumHamiltonian2024,yuEfficientPredictionSO3Equivariant2025,kimHighorderEquivariantFlow2025,kimMachineLearningHamiltonians2026}, which can then serve as an SCF initialization $\mathbf{P}^{(0)} = \mathcal{M}_{\mathbf{F}\rightarrow\mathbf{P}} (\hat{\mathbf{F}})$. \citet{liEnhancingScalabilityApplicability2025} target the scalability of Hamiltonian prediction to larger molecules and basis sets, introducing an alternative loss (WALoss).
    Models predicting \textbf{density matrices $\hat{\mathbf{P}} \approx \mathbf{P}^*$} share the same architectural blueprint, only differing in the supervision target~\citep{hazraPredictingOneParticleDensity2024,febrerGraph2MatUniversalGraph2025}.

    \looseness=-1
    \citet{liuUniversallyTransferableAcceleration2025} have shown that these matrix-based approaches fail to transfer to larger systems, leading to SCF \emph{deceleration} when used as initial guesses, even when trained with WALoss.
    To bridge this generalization gap, they suggest \textbf{predicting the coefficients $\mathbf c \in \mathbb{R}^{B_\text{aux}}$ of a linear auxiliary basis expansion}~\citep{songNeuralSCFNeuralNetwork2024} of the density
    $
        \rho(r) = \sum_\mu c_\mu \chi^{_{(\textrm{aux})}}_\mu(r) \:,
    $
    as a better learning target. They have demonstrated that this approach extrapolates well to larger molecules, achieving an RIC reduction of $33\%$ and a wall-time speedup of $23\%$ on PBE. However, the auxiliary expansion ansatz is limited to local functionals and cannot represent the density matrix required by hybrid functionals. For hybrid functionals, \citet{liuUniversallyTransferableAcceleration2025} fall back to traditional (non-ML) density matrix initialization for the HF-exchange component, resulting in a speedup of 16\% in their OOD evaluations on B3LYP. For exchange fractions close to unity (e.g., pure HF), the method is not applicable.

    Beyond ML approaches, traditional initial-guess strategies remain relevant baselines. In their systematic assessment of initial guess strategies, \citet{lehtolaAssessmentInitialGuesses2019} note that it is difficult to separate the effects of the initial guess from those of the SCF dynamics on convergence characteristics. Instead, they proposed using the projection $Q$ of the initial guess orbitals onto the converged ground state $Q(P^{(0)}, P^*) = \sum_{\mu \nu}^B \left(\mathbf{P}^{(0)} \mathbf{S} \, \mathbf{P}^{*} \mathbf{S}\right)_{\mu \nu}$ as a continuous metric. Among the traditional baselines, they found that superposition of atomic potentials (SAP), extended Hückel, and superposition of atomic densities (SAD) performed best, with SAD being the default in most quantum chemistry codes. More recently, \citet{yuAccelerationSelfConsistentField2025} evaluated basis-set projection and many-body expansion as non-ML initial guess strategies, achieving an average wall-time speedup of less than 3\% for B3LYP.

    \citet{zhangSelfConsistencyTrainingDensityFunctionalTheory2024} exploit the self-consistency condition to train Hamiltonian models without DFT labels, penalizing the Frobenius residual of $\hat{\mathbf{F}} - \mathcal{M}_{\mathbf{P}\to\mathbf{F}}(\mathcal{M}_{\mathbf{F}\to\mathbf{P}}(\hat{\mathbf{F}}))$ and differentiating through the Roothaan-Hall Eq.~\eqref{app:eq:Roothaan-Hall}.
    This residual treats all matrix entries uniformly and penalizes unphysical directions that leave the density unchanged (Appendix~\ref{app:density-optimization-gradient})~\citep{eberhardPhysicsAlignedElectronicGroundState2026}.
    \ours{} instead supervises the solver \emph{trajectory}, using an energy gradient loss term (Eq.~\ref{eq:SAIL-loss}), which directly measures distance to energy stationarity.
    \citet{zhangDifferentiableQuantumChemistry2022} avoid the initial-guess dependence via the implicit function theorem.
    We build on the differentiable SCF implementation of~\citet{gaoLearningEquivariantNonLocal2024}, which supports backpropagation through the solver and its convergence accelerator (Appendix~\ref{app:diis}), and show that differentiating through a single SCF step ($T=1$) suffices in practice.

\section{Methods}

    To study the effect of \ours{} on different prediction targets, we use the same \textbf{backbone} for all approaches. We adapt the NeuralSCF embedding~\citep{songNeuralSCFNeuralNetwork2024}, constructing SE(3)-equivariant node features from atomic numbers $Z_a$ and pairwise displacements $\mathbf{r}_{ij}$, and feed them into an EquiformerV2 GNN~\citep{liaoEquiformerV2ImprovedEquivariant2023} (hyperparameters in Appendix~\ref{app:hyperparameters}).

    We combine this backbone with \textbf{ansatz-dependent readout heads}. For the \textbf{coefficient-readout}, we follow~\citet{songNeuralSCFNeuralNetwork2024} and group the auxiliary basis coefficients $c_\nu$ by the atomic number $Z_\nu$ of their parent atom and their associated angular momentum $l_\nu$. The node-wise readout $\hat{f}_\nu$ is then elementwise rescaled as $\hat{c}_\nu = \hat{f}_\nu\,\sigma_{(Z_\nu,l_\nu)} + \mu_{(Z_\nu,l_\nu)}$, with mean $\mu$ and standard deviation $\sigma$ computed per group from the training data. Appendix~\ref{app:coeff-fock} describes in detail how these predicted auxiliary coefficients are converted into an initial density $\mathbf{P}^{(0)}$.
    For \textbf{matrix prediction}, we adopt the QHNet module~\citep{yuEfficientEquivariantGraph2023} and predict the target matrix $\mathbf{X} \in \{\mathbf{F}^{(-1)}, \mathbf{P}^{(0)}\}$ block-by-block. Each atom pair $(i,j)$ contributes a submatrix $\mathbf{X}_{ij} \in \mathbb{R}^{B_{Z_i} \times B_{Z_j}}$, built from a self-tensor product of node features (diagonal/node-prediction) or a filtered tensor product of both node features (off-diagonal/edge-prediction). Fock-prediction models additionally require the map $\mathbf{P}^{(0)} = \mathcal{M}_{\mathbf{F}\to\mathbf{P}}(\mathbf{F}^{(-1)})$ to obtain the initial density guess.

    The \textbf{baseline} models are trained on surrogate ground-state losses, which also serve as the pretraining stage for our fine-tuned \ours{} comparisons.
    We train the coefficient models by minimizing the $L^2$ density error
    \begin{align}
        \mathcal{L}_{\mathrm{coeff}}(\hat c, c^*)
        = \int \bigl|\rho^*(\mathbf{r}) - \hat{\rho}(\mathbf{r})\bigr|^2 \, d\mathbf{r}
        = \sum_{\mu \nu}^{B_\textrm{aux}} \Delta c_\mu \left(\int \chi^{(\mathrm{aux})}_\mu(\mathbf{r}) \, \chi^{(\mathrm{aux})}_\nu(\mathbf{r}) \, d\mathbf{r}\right) \Delta c_\nu \:,
    \end{align}
    where $\Delta c_\mu = \hat{c}_\mu - c^*_{\mu}$. The $B_\textrm{aux}^2$ integrals depend only on the molecular geometry and can be precomputed once per structure. The matrix-based models aim to minimize a mixed Frobenius--$L^1$ loss on the prediction error $\Delta\mathbf{X} = \hat{\mathbf{X}} - \mathbf{X}^{*}$ of a matrix target $\mathbf{X} \in \{\mathbf{P}, \mathbf{F}\}$
    \begin{align}
        \mathcal{L}_{\mathbf X}
        &= \frac{1}{2 B}\!\left(
             \sqrt{\sum_{\mu \nu} (\Delta X_{\mu \nu})^{2}}
             + \sum_{\mu \nu} \lvert\Delta X_{\mu \nu}\rvert
           \right) \:.
    \end{align}

\begin{figure}[t]
        \centering
        \vspace{0.4cm}
        \includegraphics[width=\textwidth]{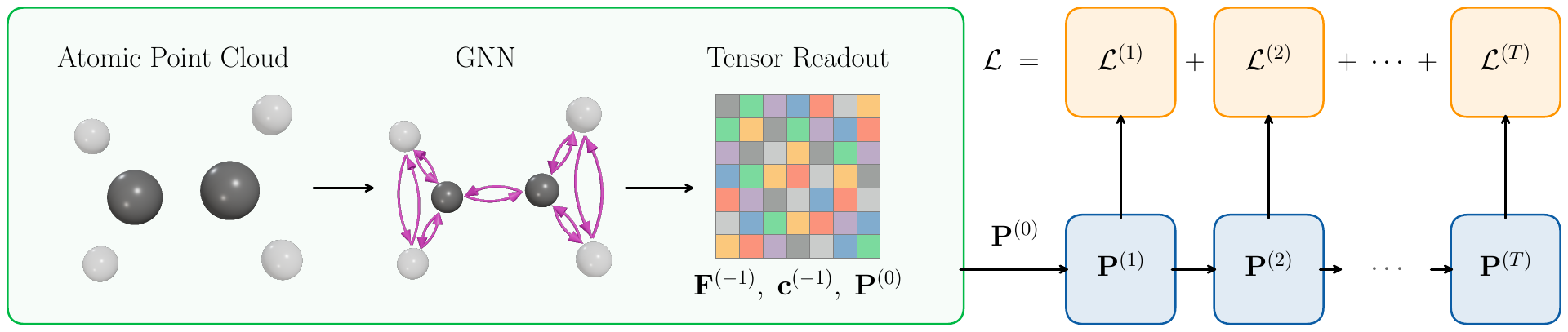}
        \caption{
              \textbf{Illustration of \oursLong{} (\ours{}).}
              ML initial-guess models use an $\text{SE}(3)$-equivariant message-passing GNN to map the point cloud of atoms $(\mathbf{Z}, \mathbf{R})$ to an initial tensor that parameterizes a basis-set expansion. If necessary, the output is converted to a density matrix, $\mathbf{P}^{(0)}$, which is used to initialize an SCF calculation. The SCF solver produces a sequence of density matrices $\mathbf{P}^{(0)}, \mathbf{P}^{(1)}, \ldots, \mathbf{P}^{(T)}$, with per-iteration losses $\mathcal{L}^{(t)}$ supervising the trajectory toward self-consistency.
          }
          \label{fig:method}
          \vspace{0.2cm}
      \end{figure}

    \textbf{\oursLong{} (\ours{})} differentiates through $T$ SCF cycles end-to-end (Fig.~\ref{fig:method}), either starting from a pretrained baseline or training directly from scratch in a single label-free stage.
    The SCF problem minimizes the total energy over the manifold of valid density matrices (Appendix~\ref{app:density-optimization-gradient}).
    At a converged solution, the energy is stationary and its gradient $\mathbf{G}^{(t)}$ vanishes~\citep{lehtolaOverviewSelfConsistentField2020}.
    To reward initial guesses that accelerate convergence, we uniformly penalize the root-mean-square gradient along the trajectory~\citep{eberhardDerivativeInformedLearning2026}
    \begin{align}\label{eq:SAIL-loss}
        \mathcal{L}_{\mathrm{tot}}
        = \frac{1}{T}\sum_{t=1}^{T} \mathcal{L}_{\nabla}^{(t)}, \qquad
        \mathcal{L}_{\nabla}^{(t)}
        = \sqrt{
            \mathbb{E}_{ia}\!\left[\,
              \bigl|G_{ia}^{(t)}\bigr|^{2}
            \right]
          }\:.
    \end{align}
    Since the energy error is second order in the density it is only a weak proxy for the distance to the ground state $\nabla E ({\mathbf{P^*}}) = \mathbf{0}$, so $\Delta E = \mathcal{O}(\|\delta\mathbf{P}\|^2)$, whereas its gradient in the manifold of valid densities is first-order $\|\mathbf{G}\| = \mathcal{O}(\|\delta\mathbf{P}\|)$.
    Additionally, $\Delta E$ collapses into a single scalar, so many different $\delta\mathbf{P}$ yield the same energy error while
    the gradient is itself an $N_{\mathrm{occ}} \times N_{\mathrm{virt}}$ matrix, so minimizing its root-mean-square puts pressure on every mode separately.
    Energy supervision is therefore informationally collapsed, whereas gradient supervision preserves the full directional structure (Appendix~\ref{app:surrogate-metrics}). Consistent with this reasoning, preliminary experiments with energy-based trajectory losses did not yield reliable SCF acceleration.

    To \textbf{measure the resulting acceleration}, prior work heavily relies on the Relative Iteration Count (RIC)~\citep{yuQH9QuantumHamiltonian2024,zhangSelfConsistencyTrainingDensityFunctionalTheory2024,febrerGraph2MatUniversalGraph2025,kimMachineLearningHamiltonians2026}
    \begin{align}
        \textrm{RIC} = \frac{\# \textrm{SCF iterations from learned initialization}}{\# \textrm{SCF iterations from } \mathbf{P^{(0)}_\textrm{ref}}}\:,
    \end{align}
    where $\mathbf{P_\textrm{ref}^{(0)}}$ is an established (non-ML) reference initialization. However, even when ignoring inference cost, RIC does not capture the relative wall-time speedup. For example, a $\Delta$-learning approach for Hamiltonian prediction requires computing $\mathbf{F}_{\mathrm{base}} = \mathcal{M}_{\mathbf{P}\rightarrow\mathbf{F}}(\mathbf{P}_{\mathrm{base}})$, adding the learned residual $\mathbf{F}^{(-1)} = \mathbf{F}_{\mathrm{base}} + \Delta\mathbf{F}_{\mathrm{learned}}$, and solving $\mathbf{P}^{(0)} = \mathcal{M}_{\mathbf{F}\rightarrow\mathbf{P}}(\mathbf{F}^{(-1)})$, a complete additional SCF cycle that does not arise for $\Delta$-learning in density matrix space. For coefficient models, we discuss the Fock build procedure in Appendix~\ref{app:coeff-fock}.
    Based on our SCF wall-time benchmarks in Figure~\ref{fig:scf-cycle}, we propose the \textbf{Effective Relative Iteration Count (ERIC)}
    \vspace{1mm}
    \begin{align}
        \label{eq:eric}
        \textrm{ERIC} = \frac{\# \textrm{ \textbf{total} Fock builds required for SCF with ML-initialization}}{\# \textrm{Fock builds when starting from } \mathbf{P^{(0)}_\textrm{ref}}}\:,
    \end{align}
    \vspace{1mm}
    which counts all Fock builds, including those hidden in the initialization. The correction is not negligible: a typical B3LYP calculation on QM9 converges in $\approx$10 cycles from $\mathbf{P}_\textrm{MINAO}$. A $\Delta$-Hamiltonian model that reduces the solver to 6 iterations achieves a 40\% RIC reduction, but the hidden Fock build yields an $\textrm{ERIC}$ of $7/10$, only 30\%.
    This gap is comparable to the RIC differences between non-$\Delta$-learned and $\Delta$-learned Hamiltonian models across the literature, including recent flow-matching approaches (Appendix~\ref{app:prior-work-reference-RIC}), suggesting that their apparent gain may largely stem from the uncounted Fock build (see Table~\ref{tab:this-work-RIC}).

\section{Experiments}
    \begin{table}[b]
        \centering
        \caption{
        \textbf{SCF acceleration of baseline and \ours{} (fine-tuned) models} across input representations and functionals, evaluated on QM40 (out-of-distribution). \textcolor{gray}{Grey rows} show ablation variants. \textbf{Bold}: best RIC/ERIC per functional across all input representations.}
        \label{tab:sail_merged}
        \small
        \begin{NiceTabular}{lll l cc cc cc}
        \CodeBefore
        \rectanglecolor{sailblue}{2-6}{\arabic{iRow}-6}
        \rectanglecolor{sailblue}{2-8}{\arabic{iRow}-8}
        \rectanglecolor{sailblue}{2-10}{\arabic{iRow}-10}
        \Body
        \toprule
        \textbf{Model} & & \textbf{Functional} & \textbf{Ablation} & \multicolumn{2}{c}{\textbf{Surrogate Loss} ($\downarrow$)} & \multicolumn{2}{c}{\textbf{RIC} ($\downarrow$)} & \multicolumn{2}{c}{\textbf{ERIC} ($\downarrow$)} \\
        & & & & base & \ours{} & base & \ours{} & base & \ours{} \\
        \midrule
        Coefficients & $\mathbf{c}$ & PBE & & $0.003$ & $0.015$ & $\mathbf{0.58}$ & $0.59$ & $0.65$ & $0.66$ \\
        & & SCAN & & $0.004$ & $17.558$ & $0.77$ & $0.67$ & $0.85$ & $0.74$ \\
        & & B3LYP & & $0.003$ & $53.815$ & $0.82$ & $\mathbf{0.68}$ & $0.90$ & $0.76$ \\
        \midrule
        Density Mat. & $\mathbf{P}$ & PBE & & $0.281$ & $1.149$ & $1.05$ & $0.60$ & $1.05$ & $\mathbf{0.60}$ \\
        & & SCAN & & $0.250$ & $1.100$ & $0.99$ & $\mathbf{0.65}$ & $0.99$ & $\mathbf{0.65}$ \\
        & & \textcolor{gray}{"} & \textcolor{gray}{T=4} & \textcolor{gray}{"} & \textcolor{gray}{$1.101$} & \textcolor{gray}{"} & \textcolor{gray}{$0.65$} & \textcolor{gray}{"} & \textcolor{gray}{$0.65$} \\
        & & \textcolor{gray}{"} & \textcolor{gray}{T=2} & \textcolor{gray}{"} & \textcolor{gray}{$1.110$} & \textcolor{gray}{"} & \textcolor{gray}{$0.65$} & \textcolor{gray}{"} & \textcolor{gray}{$0.65$} \\
        & & \textcolor{gray}{"} & \textcolor{gray}{T=1} & \textcolor{gray}{"} & \textcolor{gray}{$1.145$} & \textcolor{gray}{"} & \textcolor{gray}{$0.65$} & \textcolor{gray}{"} & \textcolor{gray}{$0.65$} \\
        & & \textcolor{gray}{"} & \textcolor{gray}{Single-stage} & \textcolor{gray}{$-$} & \textcolor{gray}{$1.270$} & \textcolor{gray}{$-$} & \textcolor{gray}{$0.66$} & \textcolor{gray}{$-$} & \textcolor{gray}{$0.66$} \\
        & & B3LYP & & $0.267$ & $0.976$ & $1.02$ & $\mathbf{0.68}$ & $1.02$ & $\mathbf{0.68}$ \\
        \midrule
        Fock & $\mathbf{F}$ & PBE & & $0.005$ & $0.277$ & $1.11$ & $0.66$ & $1.18$ & $0.73$ \\
        & & SCAN & & $0.017$ & $0.284$ & $1.16$ & $0.69$ & $1.23$ & $0.76$ \\
        & & B3LYP & & $0.026$ & $0.382$ & $1.47$ & $0.70$ & $1.55$ & $0.78$ \\
        \bottomrule
        \end{NiceTabular}
    \end{table}

    We evaluate \ours{} on QM9~\citep{ramakrishnanQuantumChemistryStructures2014}, QM40~\citep{madushankaQM40RealisticQuantum2024}, and QMugs~\citep{isertQMugsQuantumMechanical2021} for PBE, SCAN, and B3LYP, representing a GGA, a meta-GGA, and a hybrid functional~\citep{perdewGeneralizedGradientApproximation1996,sunStronglyConstrainedAppropriately2015,beckeDensityfunctionalThermochemistryIII1993,perdewJacobsLadderDensity2001}. We split QM9 by molecular size (train: $\leq20$, val: $21$--$22$, test: $\geq23$ atoms). To evaluate size extrapolation beyond QM9, we use QM40 and QMugs as far-out-of-distribution test sets. For a head-to-head comparison of the coefficient-based models proposed by~\citet{liuUniversallyTransferableAcceleration2025}, we evaluate their performance when trained directly on semi-local and hybrid functionals using the authors' proposed extensions.
    For additional evaluation details, see Appendix~\ref{app:evaluation details}. We compare to prior ML methods in Appendix~\ref{app:prior-work-reference-RIC} and to traditional functional-initialization baselines in Appendix~\ref{app:traditional baselines}.

    Table~\ref{tab:sail_merged} reports our \textbf{far-out-of-distribution size-extrapolation} results on QM40, together with key \textbf{design ablations}. Since \ours{} fine-tuning starts from the baseline parameters, we can directly compare surrogate losses before and after fine-tuning. The matrix-based baselines decelerate SCF convergence out of distribution (ERIC $>1$), confirming the failure mode reported by~\citet{liuUniversallyTransferableAcceleration2025}.
    \textbf{The baseline failures are due to misalignment, not extrapolation.} The surrogate losses in Table~\ref{tab:sail_merged} stay small out of distribution, so the baselines \emph{do} extrapolate, just not the right quantity. Matrix models overfit ground-state targets in directions that decelerate the solver, which is why \textbf{\ours{} restores acceleration (ERIC $<1$) for both matrix-based ansätzes across all three functionals}.
    The coefficient model separates into two categories. On PBE, it already extrapolates after ground-state training (ERIC $0.65$) and shows only minor misalignment, suggesting that the restricted expressivity of the linear auxiliary expansion acts as an implicit regularization. On SCAN and B3LYP, extending the coefficient ansatz beyond GGA requires Fock-build substitutions (Appendix~\ref{app:coeff-fock}), since the density alone does not determine the kinetic-energy density $\tau$ or the exchange matrix $\mathbf{K}$. Because the ansatz is misspecified, the ground-state coefficient target is a poor learning objective (degrading the baseline to ERIC $0.85$ and $0.90$, respectively), whereas \ours{} lets the model absorb these systematic Fock-build errors into a compensating adjustment that drives the auxiliary surrogate loss up by orders of magnitude but the ERIC down.

\begin{figure}[t]
        \centering
        \includegraphics[width=\textwidth]{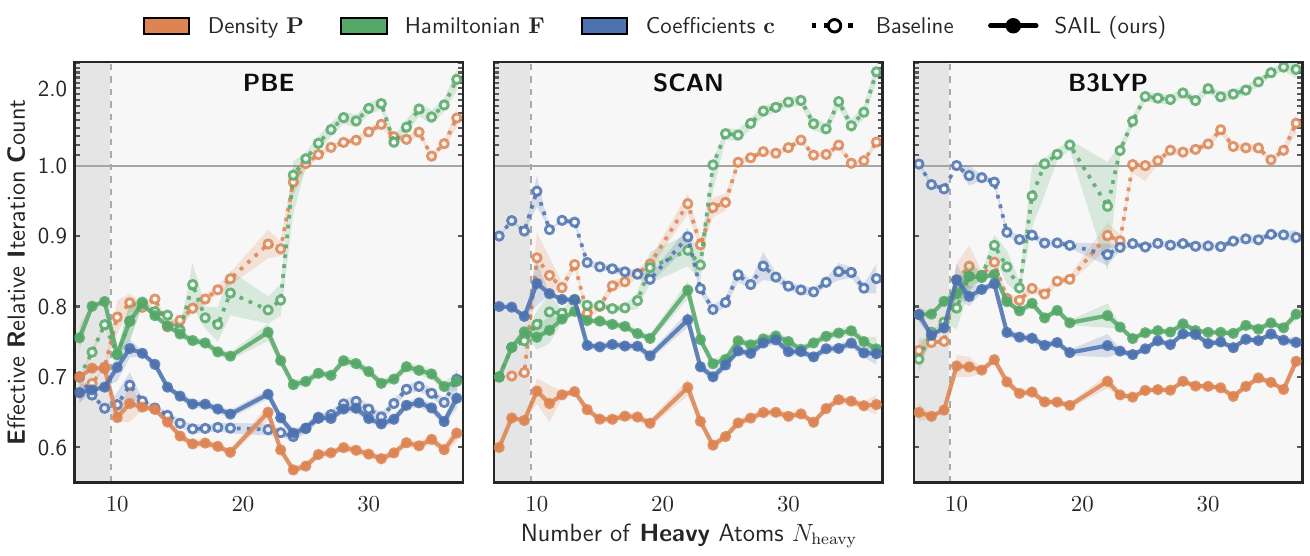}
        \vspace{-0.5cm}
        \caption{
            \textbf{Effective Relative Iteration Count (ERIC)} for different XC-functionals (from left to right: local, semi-local, hybrid) on far-out-of-distribution molecules, binned by their number of heavy atoms (up to 71 atoms, 37 heavy). All models are \textbf{trained on QM9} molecules with at most 20 atoms; we only plot QM9 test molecules with $>23$ atoms. The dashed vertical line marks the transition from the QM9 test set to the QM40 test set. For comparison to related work, see Table~\ref{tab:prior-work-RIC}.
          }
        \label{fig:eric}
        \vspace{-0.1cm}
    \end{figure}

\begin{figure}[t]
          \centering
          \vspace{-0.1cm}
          \includegraphics[width=\textwidth]{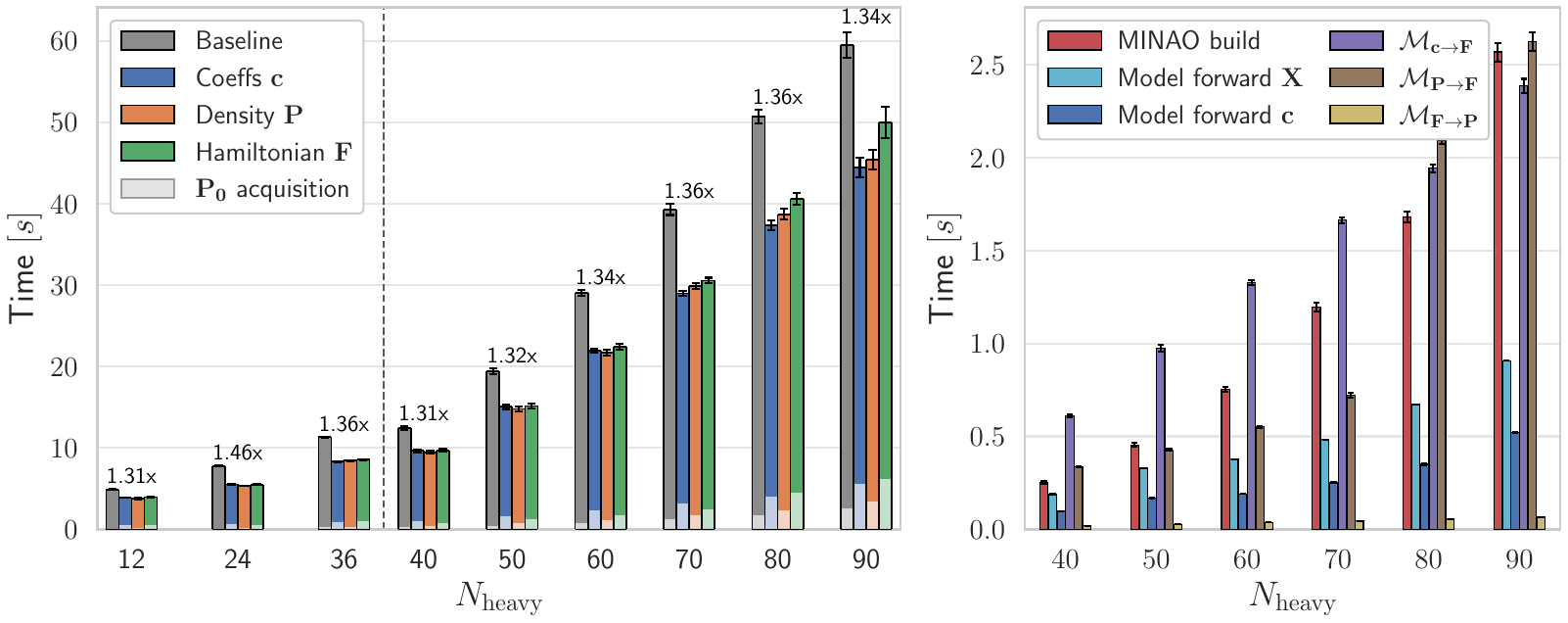}
          \vspace{-0.6cm}
          \caption{%
            \textbf{Left: Wall-time speedup on B3LYP/def2-SVP.}
            Total SCF wall-time on QM40 (left of dashed line) and QMugs, measured on GPU with GPU4PySCF. Each bar is decomposed into initial-guess acquisition (lighter color) and the subsequent SCF loop. All \ours{}-trained models deliver consistent speedups across the full size range, with no degradation on molecules $10\times$ larger than the training distribution. Numbers above the bars report the speedup of the density matrix model, which outperforms the Hamiltonian model at every size.
            \textbf{Right: Initial-guess acquisition breakdown.}
            Per-component cost of the initial-guess acquisition on QMugs. Matrix-based models use $\Delta$-learning on top of MINAO, so the MINAO build is a fixed cost shared by both ansätzes. The model forward pass adds only a small overhead, while the extra Fock build $\mathcal{M}_{\mathbf{P}\to\mathbf{F}}$ required by $\Delta$-Hamiltonian learning dominates at large sizes, motivating ERIC.
            }
            \vspace{-0.3cm}
          \label{fig:b3lyp}
    \end{figure}

    Figure~\ref{fig:eric} shows ERIC as a function of heavy-atom count on far-out-of-distribution molecules from QM40. Without \ours{}, matrix-based models degrade with increasing molecule size, confirming the failure mode reported by~\citet{liuUniversallyTransferableAcceleration2025}.  \textbf{After \ours{} fine-tuning, the ERIC remains flat across all three ansätzes and functionals, with no upward trend even at four times the training-distribution size} (up to 9 heavy vs. up to 37 heavy).
    The coefficient-based model also benefits from \ours{} on SCAN and B3LYP and is on par on PBE (Figure~\ref{fig:eric}).
    For B3LYP, we extend our size extrapolation experiment to QMugs (Figure~\ref{fig:Qmugs-B3LYP-eric}). \ours{} maintains a mean ERIC of $0.74$ (density matrix prediction) with \textbf{no upward trend even for molecules $10\times$ larger than the training distribution}.
    \textbf{The B3LYP wall-time measurements} for matrix-based methods (Figure~\ref{fig:b3lyp}) show that these \textbf{ERIC reductions translate into comparable speedups}, as expected from the Fock-build dominance at the hybrid level (Figure~\ref{fig:scf-cycle}), while the per-component breakdown shows that model evaluation is cheap compared to the MINAO initial-guess cost.
    Additional B3LYP runs with three training seeds show that the \ours{} results are reproducible, with seed-to-seed variations of at most $0.01$ ERIC, and \ours{}- and MINAO-initialized calculations reach the same final energies to within $10^{-7}\:\textrm{Ha}$ (Appendix~\ref{app:robustness}).

    \begin{wraptable}{r}{0.42\textwidth}
        \centering
        \vspace{-0.7cm}
        \caption{
            \textbf{Per-cycle parameter gradients} $\mathbf{g}_t = \nabla_{\boldsymbol{\theta}} \mathcal{L}_{\nabla}^{(t)}$ for SCAN density-matrix \ours{} on QM40 ($60$ molecules).
        }
        \label{tab:gradient-collinearity}
        \vspace{0.15cm}
        \small
        \setlength{\tabcolsep}{4pt}
        \begin{tabular}{l cccc}
            \toprule
            \textbf{Cycle} $t$ & $1$ & $2$ & $4$ & $8$ \\
            \midrule
            $\|\mathbf{g}_t\|$ & \texttt{7e-3} & \texttt{5e-3} & \texttt{2e-4} & \texttt{2e-8} \\
            $\cos(\mathbf{g}_t, \mathbf{g}_1)$\hspace{0.2cm} & $1.00$ & $1.00$ & $0.98$ & $0.96$ \\
            \bottomrule
        \end{tabular}
        \vspace{-0.4cm}
    \end{wraptable}
    \looseness=-1
    To test whether \ours{} requires unrolling the full SCF trajectory or pretraining on ground-state labels, we ablate both components on the SCAN density-matrix model in Table~\ref{tab:sail_merged}.
    \textbf{Truncating backpropagation to a single solver cycle ($T=1$) matches the ten-cycle default} on QM40 at an ERIC of $0.65$, as do $T=2$ and $T=4$.
    At $T=1$, the gradient still passes through the first SCF update (two Fock builds and one eigensolve) and omits only deeper cycles where DIIS extrapolation acts.
    Table~\ref{tab:gradient-collinearity} explains why deeper unrolling adds little.
    Under DIIS, the per-cycle parameter-gradient norm $\|\mathbf{g}_t\|$ decays by five orders of magnitude by cycle $8$, while the direction $\cos(\mathbf{g}_t, \mathbf{g}_1)$ stays aligned with the first cycle across all steps.
    \textbf{Training directly from scratch without ground-state pretraining} (single-stage) reaches an ERIC of $0.66$, matching the pretrained model ($0.65$) and improving over the surrogate-trained baseline ($0.99$) without any reference DFT labels. Additional ablations of model design choices and training objectives are provided in Appendix~\ref{app:ablations}.

    \begin{wraptable}{r}{0.42\textwidth}
        \centering
        \vspace{-0.7cm}
        \caption{
            \textbf{Simultaneous size and basis set generalization} for coeff. models on B3LYP.
            Trained on QM9, eval. on QM40.
        }
        \vspace{0.15cm}
        \label{tab:basis-set-generalization}
        \small
        \setlength{\tabcolsep}{4pt}
        \begin{NiceTabular}{l cc cc}
            \CodeBefore
            \rectanglecolor{sailblue}{2-3}{\arabic{iRow}-3}
            \rectanglecolor{sailblue}{2-5}{\arabic{iRow}-5}
            \Body
            \toprule
            \textbf{Basis Set} & \multicolumn{2}{c}{\textbf{RIC} ($\downarrow$)} & \multicolumn{2}{c}{\textbf{ERIC} ($\downarrow$)} \\
                        & base   & \ours{} & base   & \ours{} \\
            \midrule
            def2-SVP (ID)    & $0.82$ & $0.68$  & $0.90$ & $0.76$  \\
            \midrule
            def2-TZVP   & $0.82$ & $\mathbf{0.74}$  & $0.90$ & $\mathbf{0.82}$  \\
            def2-TZVPPD & $0.82$ & $\mathbf{0.74}$  & $0.90$ & $\mathbf{0.82}$  \\
            def2-QZVP   & $0.82$ & $\mathbf{0.74}$  & $0.90$ & $\mathbf{0.82}$  \\
            \bottomrule
        \end{NiceTabular}
        \vspace{-0.4cm}
    \end{wraptable}
    \looseness=-1
    \textbf{Simultaneous size and basis set generalization.} Matrix ansätzes tie predictions to a specific AO basis and require retraining for each. Coefficient models predict in a fixed auxiliary basis and apply unchanged to any compatible target AO basis~\citep{liuUniversallyTransferableAcceleration2025}. Table~\ref{tab:basis-set-generalization} combines size and basis extrapolation. We train on QM9/def2-SVP and evaluate on QM40 using larger basis sets. The baseline coefficient model stays near ERIC $0.90$ across the four bases. \ours{} reduces ERIC further to $0.82$ on the three larger bases. The \ours{} gain is smaller outside the training basis set ($8$ vs. $14$ percentage points of the $\mathbf{P}_{\mathrm{MINAO}}$ cost), but \ours{} outperforms the baseline at every target.

    \textbf{To test generalization to non-equilibrium geometries} we evaluate the robustness of acceleration
    under geometric distortions for QM9-trained models on 3BPA~\citep{kovacs2021linear}.
    This benchmark contains conformations of 3-(benzyloxy)pyridin-2-amine sampled at 300, 600, and 1200\,K, together with angular dihedral scans.
    As temperature increases from 300 to 1200\,K, its mean ERIC rises modestly from $0.81$ to $0.86$, whereas the surrogate-trained baseline deteriorates, becoming slower than MINAO.
    Table~\ref{tab:3bpa} shows that \textbf{the density-matrix model retains acceleration across all tested temperatures and dihedral scans}.
    On the dihedral scans, \ours{} achieves ERICs of $0.77$--$0.78$, compared with $0.89$--$0.90$ for the baseline.
    \begin{wraptable}{r}{0.44\textwidth}
        \centering
        \vspace{-0.5cm}
        \caption{
            \textbf{Transfer to non-equilibrium geometries on 3BPA.}
            ERIC at B3LYP for models trained on QM9.
        }
        \label{tab:3bpa}
        \vspace{0.15cm}
        \small
        \begin{NiceTabular}{lcccc}
            \CodeBefore
            \rectanglecolor{sailblue}{2-3}{\arabic{iRow}-3}
            \rectanglecolor{sailblue}{2-5}{\arabic{iRow}-5}
            \Body
            \toprule
            \textbf{Test set} & \multicolumn{2}{c}{\textbf{Density $\mathbf{P}$}}
            & \multicolumn{2}{c}{\textbf{Coefficients $\mathbf{c}$}} \\
            & Base & \ours{} & Base & \ours{} \\
            \midrule
            300\,K                 & 0.94 & \textbf{0.81} & 0.94 & 0.86 \\
            600\,K                 & 0.99 & \textbf{0.84} & 0.96 & 0.91 \\
            1200\,K                & 1.08 & \textbf{0.86} & 1.00 & 0.99 \\
            \midrule
            $\beta=120^\circ$ & 0.89 & \textbf{0.77} & 0.95 & 0.86 \\
            $\beta=150^\circ$ & 0.90 & \textbf{0.78} & 0.95 & 0.87 \\
            $\beta=180^\circ$ & 0.90 & \textbf{0.78} & 0.96 & 0.87 \\
            \bottomrule
        \end{NiceTabular}
        \vspace{-0.4cm}
    \end{wraptable}
    The coefficient model also benefits from \ours{}, but its advantage diminishes with temperature, reaching ERIC $0.99$ at 1200\,K. This contrast supports density-matrix prediction as the more robust choice in these tests.
    We further evaluate transfer to reactive pathways and transition states from Transition1x~\citep{schreinerTransition1xDatasetBuilding2022} in Appendix~\ref{app:transition-states} (Tables~\ref{tab:transition-states-h-transfer} and~\ref{tab:transition-states-dissociation}). \ours{} accelerates all frames along an intramolecular hydrogen transfer and through the transition state of a $\mathrm{C{-}C}$ bond dissociation, whereas compressed contact dimers past bond breaking are not accelerated ($\mathrm{ERIC}=1.00$, likely because mid-dissociation geometries are absent from the training set), though never decelerated.

    \begin{figure}[t]
      \centering
      \includegraphics[width=\textwidth]{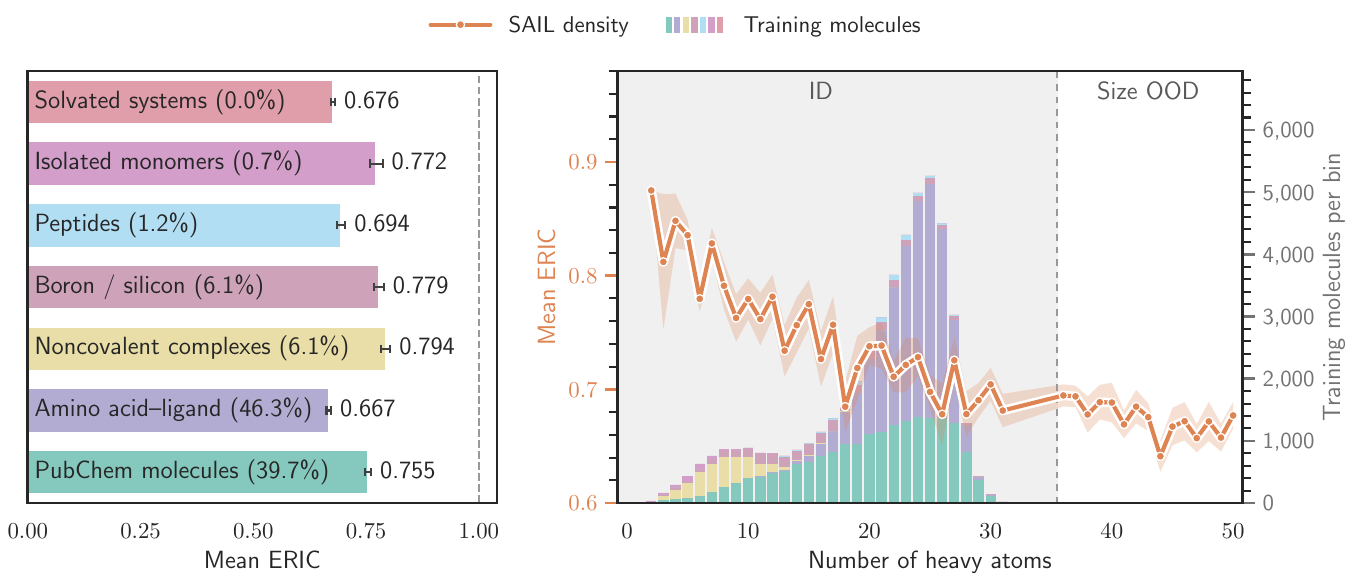}
      \vspace{-0.4cm}
      \caption{
        \textbf{Chemical coverage and size transfer on SPICEv2 for hybrid-level DFT.}
        We train a density-matrix model label-free from scratch using the minimal, $T=1$, version of \ours{} without surrogate pretraining.
        \textbf{Left:} Mean ERIC across chemical categories; percentages indicate each category's share of the 48,470 training molecules.
        \textbf{Right:} Mean ERIC versus heavy-atom count (orange, left axis), overlaid on the number of training samples per bin (gray, right axis).
        }
      \label{fig:SPICEv2-B3LYP-eric}
    \end{figure}
    All models up to here have been trained on QM9, which contains only neutral equilibrium molecules across five elements (H, C, N, O, F).
    To show-case the \textbf{robustness of \ours{} in a more chemically diverse setting while minimizing up-front costs},
    we directly train a density-matrix model on small molecules from SPICEv2~\citep{eastman2024nutmeg}
    \textbf{without pretraining (label-free) for only two epochs, differentiating through a single solver cycle ($T=1$)}.
    The training set consists of $48,470$ structures at B3LYP / def2-SVP level of theory and spans neutral and charged systems across 13 elements (H, B, C, N, O, F, Na, Si, P, S, Cl, K, and Br).
    Despite this short, single-stage, one-step training, Figure~\ref{fig:SPICEv2-B3LYP-eric} shows that \textbf{\ours{} accelerates every evaluated chemical category while also extrapolating to larger molecules}.
    Across all subsets, mean ERIC stays between $0.67$ and $0.79$ regardless of training share, including zero-shot transfer to solvated systems absent from the training set ($0.68$).
    In the size-extrapolation regime, mean ERIC \emph{improves} from $0.75$ in distribution ($\leq 35$ heavy atoms) to $0.67$ in the size-extrapolation regime ($36$-$50$ heavy atoms).
    Across all 600 evaluation structures (300 ID and 300 OOD), 599 converge under both initializations, and \textbf{\ours{} never decelerates a single calculation}, reducing iterations on 596 structures (including all 300 OOD structures), matching MINAO on three, and failing alongside MINAO on one H/N/Si structure.

\section{Discussion}
    \label{sec:discussion}
    This work introduces \textbf{\oursLong{} (\ours{})}, which trains on solver dynamics rather than ground-state targets, and the \textbf{Effective Relative Iteration Count (ERIC)}, which corrects RIC for ansatz-dependent Fock-build overhead.
    We evaluate ML initial-guess architectures across three rungs of Jacob's ladder, a GGA (PBE), a meta-GGA (SCAN), and a hybrid (B3LYP).

    On QM9/QM40, we independently confirm the size-extrapolation failure of conventionally trained matrix-based methods reported by~\citet{liuUniversallyTransferableAcceleration2025}.
    We show that this failure stems from a \textbf{misalignment between ground-state supervision and solver dynamics}. Models trained to minimize a loss with respect to the converged solution do not produce initial guesses that converge fast.
    \ours{} \textbf{cures the size-extrapolation failure for matrix methods}, achieving near-identical ERICs on QM40 ($\mathbf c$: $0.66$, $\mathbf P$: $0.60$, $\mathbf F$: $0.73$, Tab.~\ref{tab:sail_merged}) at the PBE/def2-SVP level of theory.

    \looseness=-1
    \textbf{\ours{} extends reliable SCF acceleration to mGGA and hybrid functionals}.
    The coefficient and $\Delta$-density matrix ansätzes offer complementary strengths in this regime.
    The coefficient approach transfers across AO basis sets without retraining (Table~\ref{tab:basis-set-generalization}).
    It does not apply to pure Hartree--Fock or to functionals with exact-exchange fractions near unity~\citep{liuUniversallyTransferableAcceleration2025}.
    For hybrid functionals, the coefficient ansatz relies on the model to compensate for the fixed exact-exchange treatment by learning a density distortion.
    Whether this also works on \emph{range-separated} hybrids remains untested.
    $\Delta$-density matrix prediction requires retraining per basis set. It uses no functional-dependent substitutions and gives the most consistent acceleration across our evaluations, reaching ERIC $0.65$ on SCAN ($35\%$ reduction over MINAO) and $0.68$ on B3LYP ($32\%$).
    \ours{} thus extends reliable ML SCF acceleration to drug-like molecules up to $10\times$ larger than the training distribution at the hybrid level of theory.

    \begin{wraptable}{r}{0.4\textwidth}
        \centering
        \vspace{-0.7cm}
        \caption{
            \textbf{NVIDIA-A100 hours saved} per $10{,}000$ B3LYP / def2-SVP calculations for \ours{}-trained $\mathbf P$ models.
        }
        \label{tab:breakeven}
        \vspace{0.2cm}
        \small
        \setlength{\tabcolsep}{4pt}
        \begin{tabular}{l cccccc}
            \toprule
            $N_{\mathrm{heavy}}$ & $40$ & $50$ & $60$ & $70$ & $80$ & $90$ \\
            \midrule
            Hours saved & $8$ & $13$ & $20$ & $29$ & $37$ & $42$ \\
            \bottomrule
        \end{tabular}
        \vspace{-0.4cm}
    \end{wraptable}
    \textbf{The training cost of \ours{} amortizes quickly} in the label-free, single-stage $T=1$ setting,
    since it does not incur the up front labeling cost of surrogate-loss pretraining.
    On QM9, a $T=1$ step takes $86\,$ms per molecule of pure compute on an H200 GPU ($14$ GPU-hours for five epochs over ${\sim}100\text{k}$ structures), and validation curves reach over $90\%$ of the final acceleration within the first epoch (${\sim}3$ GPU-hours).
    On SPICEv2, two epochs over $48{,}470$ molecular groups take $330$--$370\,$ms of compute per step (${\sim}9$ GPU-hours of compute, or $1.7$ days wall-clock including dataloader I/O).
    In downstream B3LYP evaluations (Table~\ref{tab:breakeven}), \ours{} saves $8$ to $42$ A100 GPU-hours per $10{,}000$ calculations across molecules of $40$ to $90$ heavy atoms, so the training investment breaks even after a few thousand to a few tens of thousands of large-molecule calculations, a small fraction of a single screening or MLIP dataset generation campaign (e.g., $665\text{k}$ structures in QMugs).

    \textbf{Limitations.}
    Like the baseline methods, SAIL requires a separate model for each functional. Matrix-based ansätzes require retraining per basis set. These costs are amortized over all subsequent calculations at that level of theory. We evaluate three functionals (PBE, SCAN, B3LYP) spanning three rungs of Jacob's ladder. Transfer to range-separated and double hybrids, as well as periodic structures~\citep{liDeeplearningDensityFunctional2022,gongGeneralFrameworkE3equivariant2023,tangDeepEquivariantNeural2024}, remains untested. Chemical coverage extends to 13 elements up to bromine via SPICEv2, but transition metals, heavier elements, and open-shell systems are not yet covered.
    Acceleration transfers to non-equilibrium geometries (3BPA, Transition1x), yet compressed contact pairs past bond breaking are not accelerated (ERIC~$=1.00$), suggesting that such geometries need to be represented in training.
    While SAIL itself is label-free, our strongest results combine it with surrogate-loss pretraining. SAIL training increases total training time by a factor of ${\sim}1.7\times$ relative to the baseline, though our ablations suggest that truncating the number of SCF steps during training can reduce this overhead substantially.

    ERIC improves on RIC as a wall-time proxy but does not replace wall-time measurements. Wall-time depends on hardware, XC-functional, basis set, and implementation. It does not scale linearly with iteration count, as different algorithms and molecule sizes shift the computational bottleneck~\citep{liuUniversallyTransferableAcceleration2025}. ML research code rarely matches the optimization of production DFT codes, making absolute wall-time comparisons across methods unreliable. ERIC partially bridges this gap by providing a hardware- and implementation-independent proxy.

    \textbf{Future work.}
    Since \ours{} is label-free, data augmentation through chemical perturbations such as H$\leftrightarrow$F substitutions and random atom position perturbations comes at no additional labeling cost.
    Adapting \ours{} to periodic systems and to alternative solvers, such as direct energy minimization and second-order SCF methods~\citep{sunGeneralSecondOrder2017}, is a direct extension of the framework presented here.
    Finally, initializing the GNN backbone from a pretrained force-field foundation model could be a path toward better generalization across chemical space at lower per-functional training cost.

\phantomsection
\addcontentsline{toc}{section}{Broader Impact}
\section*{Broader Impact}
    \oursLong{} accelerates an established numerical method (KS-DFT) without altering the underlying physics or accuracy, so it inherits rather than introduces risk profiles relative to conventional DFT. The expected positive impacts are reduced compute and energy cost per calculation at constant accuracy, as well as faster research iteration in domains that rely heavily on DFT, including drug discovery and the generation of reference data for machine-learned interatomic potentials~\citep{kulichenkoDataGenerationMachine2024a}. On the negative side, cheaper computations may induce a rebound effect, in which aggregate energy consumption rises because lower per-calculation costs expand the volume of calculations performed. We are not aware of direct dual-use concerns specific to SCF acceleration beyond those already present for DFT itself.

\phantomsection
\addcontentsline{toc}{section}{Acknowledgments}
\section*{Acknowledgments}
    We thank Zhe Liu, Siyuan Liu, Qiming Sun, Arghya Bhowmik, and Pol Febrer for insightful discussions regarding their respective methods, and Nicholas Gao for feedback on the final manuscript.

\bibliographystyle{plainnat}
\renewcommand{\bibsection}{%
  \phantomsection
  \addcontentsline{toc}{section}{\refname}%
  \section*{\refname}%
}
\bibliography{references}

\newpage
\appendix

\startcontents[appendix]
\printcontents[appendix]{}{1}{%
\section*{Contents}
\setcounter{tocdepth}{2}%
}
\clearpage

\section{The Self-Consistent Field Cycle}
    \label{app:scf-details}
    For accessibility, this section is limited to the spin-restricted Kohn-Sham (RKS) framework.
    In practice, the electron density $\rho$ is expanded in a finite basis set $\{\chi_\mu\}_{\mu=1}^B$ of atom-centered functions via a coefficient matrix $\mathbf{C} \in \BxB$,
    \begin{align}
        \rho(\mathbf{r}) &= 2  \sum_{\mu \nu}^B \sum_{i=1}^{N_e/2} C_{\mu i} \, C_{\nu i}
        \, \chi_\mu(\mathbf{r}) \, \chi_\nu(\mathbf{r})
        =  \sum_{\mu \nu}^B P_{\mu\nu} \, \chi_\mu(\mathbf{r}) \, \chi_\nu(\mathbf{r}) \:,
    \end{align}
    where $P_{\mu\nu} = 2 \sum_{i=1}^{N_{\mathrm{occ}}} C_{\mu i} \, C_{\nu i}$ is the density matrix and the sum runs over the $N_{\mathrm{occ}} = N_e/2$ lowest-energy \emph{occupied} orbitals, with the remaining $N_{\mathrm{virt}} = B - N_{\mathrm{occ}}$ columns of $\mathbf{C}$ termed \emph{virtual} orbitals. In its basis-set discretization, the ground-state energy becomes a \emph{function} of the density matrix
    \begin{align}
        E(\mathbf{P}) = \sum_{\mu \nu} H^{\text{(core)}}_{\mu\nu} P_{\mu\nu}
        + \tfrac{1}{2} \sum_{\mu \nu} J_{\mu\nu}(\mathbf{P}) P_{\mu\nu}
        + E_{\text{xc}}(\mathbf{P}) \:,
    \end{align}
    where $\mathbf{H}^{\text{core}}$ contains kinetic and nuclear attraction terms, $\mathbf{J}$ is the Coulomb matrix, and $E_{\text{xc}}$ is the exchange-correlation energy. The Fock matrix $\mathbf F \in \BxB$ is defined as the derivative of the energy w.r.t the density matrix
    \begin{align}
        \mathbf{F}(\mathbf{P}) = \frac{\partial E}{\partial \mathbf{P}}
        = \mathbf{H}^{\text{(core)}} + \mathbf{J}(\mathbf{P})
        + \mathbf{V}^{\text{xc}}(\mathbf{P}) \:.
    \end{align}
    At a minimum of $E(\mathbf{P})$, the coefficients $\mathbf{C}$ satisfy the Roothaan--Hall equations,
    \begin{align}
        \label{app:eq:Roothaan-Hall}
        \left(\mathbf F(\mathbf{P}) \, \mathbf C\right)_{\mu i}
        = \varepsilon_i \, \left(\mathbf S \, \mathbf C\right)_{\mu i} \:,
    \end{align}
    a generalized eigenvalue problem, where $\varepsilon_j$ are the so-called orbital energies~\citep{roothaanNewDevelopmentsMolecular1951}.
    While the overlap matrix $S_{\mu\nu} = \int \chi_\mu(\mathbf{r}) \, \chi_\nu(\mathbf{r}) \, d\mathbf{r}$ is constant with respect to $\mathbf{C}$, the Fock matrix depends on $\mathbf{C}$ through $\mathbf{P}$, making Eq.~\eqref{app:eq:Roothaan-Hall} nonlinear.
    The self-consistent field ansatz~\citep{lehtolaOverviewSelfConsistentField2020} linearizes Eq.~\eqref{app:eq:Roothaan-Hall} via the iterative solution approach
    \begin{align}
        \label{app:eq:scf-detailed}
        &\mathbf P^{(t)} \hspace{4mm}= \mathbf P \big(\mathbf C^{(t)}\big) \nonumber \\
        &\mathbf F^{(t)} \hspace{4mm}= \mathbf F \big(\mathbf P^{(t)}\big) \\
        &\mathbf C^{(t+1)} = \text{gEVP}\big(\mathbf F^{(t)}, \mathbf{S} \big) \nonumber \:,
    \end{align}
    where $\text{gEVP}(\cdot,\cdot)$ returns the eigenvectors of the generalized eigenvalue problem and $t$ is the iteration index (Fig.~\ref{fig:scf-cycle}).

    \subsection{Direct Inversion of the Iterative Subspace (DIIS)}
        \label{app:diis}
        The self-consistent field (SCF) method uses an iterative ansatz to solve Equation~\eqref{app:eq:Roothaan-Hall}.
        The so-called Direct Inversion of the Iterative Subspace (DIIS) formalism accelerates convergence by constructing an optimal linear combination of previous Fock matrices that minimizes a residual error~\citep{pulayConvergenceAccelerationIterative1980,pulayImprovedSCFConvergence1982}.
        At SCF iteration $t$, we compute the residual $\mathbf R^{(t)} \in \BxB$ (commutator error):
        \begin{align}
            \mathbf R^{(t)} = \mathbf F^{(t)} \mathbf P^{(t)} \mathbf S - \mathbf S \mathbf P^{(t)} \mathbf F^{(t)}
        \end{align}
        where $\mathbf{F}$ is the Fock matrix, $\mathbf{P}$ is the density matrix, and $\mathbf{S}$ is the basis overlap matrix.
        In practice, one often transforms the residual into an orthonormal basis via $\tilde{\mathbf{R}}^{(t)} = \mathbf{X}^\top \mathbf{R}^{(t)} \mathbf{X}$ with $\mathbf{X} = \mathbf{S}^{-1/2}$, which improves numerical stability.
        Using the DIIS method, we compute
        \begin{align}
            \mathbf{F}_{\text{DIIS}}^{(t)} = \sum_{i=1}^{t} a^{(i)} \mathbf{F}^{(i)},
        \end{align}
        to construct the next Fock matrix for which the gEVP is solved.
        The coefficients $\mathbf a^{(t)}$ minimize the residual norm subject to the constraint $\sum^t_{i=1} a^{(i)} = 1$:
        \begin{align}
            \min_{\mathbf{a} \in \mathbb{R}^t} \sum_{jk}^{t} \sum_{\mu \nu}^B a^{(j)} a^{(k)} \, R^{(j)}_{\mu \nu} \, R^{(k)}_{\mu \nu} \:.
        \end{align}
        Using Lagrange multipliers, one can derive the Pulay equations~\citep{pulayConvergenceAccelerationIterative1980}
        \begin{align}
            \begin{pmatrix}
                B_{11} & \cdots & B_{1t} & -1 \\
                \vdots & \ddots & \vdots & \vdots \\
                B_{t1} & \cdots & B_{tt} & -1 \\
                -1 & \cdots & -1 & 0
                \end{pmatrix}
                \begin{pmatrix}
                a^{(1)} \\ \vdots \\ a^{(t)} \\ \lambda
                \end{pmatrix}
                =
                \begin{pmatrix}
                0 \\ \vdots \\ 0 \\ -1
            \end{pmatrix}
        \end{align}
        where $B_{jk} := \sum_{\mu \nu} R^{(j)}_{\mu \nu} \, R^{(k)}_{\mu \nu}$.

    \subsection{Energy Gradient on the Density Matrix Manifold}
        \label{app:density-optimization-gradient}

        The SCF fixed-point iteration (Eq.~\ref{app:eq:orbital-gradient}) is not the only approach
        to finding the ground-state density matrix. An alternative is \emph{direct minimization}:
        treating $E(\mathbf{P})$ as an objective function and optimizing it with gradient-based methods
        on the constraint set of valid density matrices~\citep{lehtolaOverviewSelfConsistentField2020}. This perspective motivates the loss function
        used in \ours{}.

        The set of valid density matrices of rank $N_{\mathrm{occ}} = N_e/2$ forms a smooth manifold
        (the Grassmannian). Its tangent space at any point is spanned by rotations that mix the
        $N_{\mathrm{occ}}$ occupied with the $N_{\mathrm{virt}} = B - N_{\mathrm{occ}}$ virtual
        columns of the coefficient matrix $\mathbf{C}^{(t)} \in \BxB$.
        Rotations within the occupied or virtual subspaces leave the density matrix unchanged and
        are therefore not degrees of freedom of the optimization.
        The energy gradient with respect to these occupied--virtual rotations is the
        $N_{\mathrm{occ}} \times N_{\mathrm{virt}}$ matrix
        \begin{align}\label{app:eq:orbital-gradient}
          G_{ia}^{(t)}
          = \left(\mathbf{C}^{\mathrm{occ},(t)\top}\,
            \mathbf{F}^{(t)}\,
            \mathbf{C}^{\mathrm{virt},(t)}\right)_{ia},
        \end{align}
        where $\mathbf{C}^{\mathrm{occ},(t)} \in \mathbb{R}^{B \times N_{\mathrm{occ}}}$
        and $\mathbf{C}^{\mathrm{virt},(t)} \in \mathbb{R}^{B \times N_{\mathrm{virt}}}$
        are the occupied and virtual blocks of $\mathbf{C}^{(t)}$,
        with $i$ indexing occupied and $a$ virtual orbitals.
        At a converged SCF solution the Fock matrix is diagonal in its own eigenbasis
        (Eq.~\eqref{app:eq:Roothaan-Hall}), so the off-diagonal block
        $\mathbf{C}^{\mathrm{occ},\top} \mathbf{F}\, \mathbf{C}^{\mathrm{virt}}$ vanishes ---
        the standard first-order optimality condition $\mathbf{G} = \mathbf{0}$.

        This gives a natural convergence measure: any iterate with large $\lVert\mathbf{G}^{(t)}\rVert$
        is far from a stationary point, regardless of how close $\mathbf{P}^{(t)}$ is to $\mathbf{P}^{*}$
        in Frobenius norm. The per-cycle loss used in \ours{} (Eq.~\eqref{eq:SAIL-loss}) is the
        root-mean-square of this gradient,
        \begin{align}
          \mathcal{L}_{\nabla}^{(t)}
          = \sqrt{
              \frac{1}{N_{\mathrm{occ}}\, N_{\mathrm{virt}}}
              \sum_{i=1}^{N_{\mathrm{occ}}} \sum_{a=1}^{N_{\mathrm{virt}}}
              \left| G_{ia}^{(t)} \right|^2
            }.
        \end{align}
        Unlike the DIIS commutator residual $\mathbf{R}^{(t)} = \mathbf{F}^{(t)}\mathbf{P}^{(t)}\mathbf{S}
        - \mathbf{S}\mathbf{P}^{(t)}\mathbf{F}^{(t)}$ (Appendix~\ref{app:diis}),
        which lives in the full $B \times B$ AO space and mixes all orbital pairs,
        $\mathbf{G}^{(t)}$ isolates the $N_{\mathrm{occ}} \times N_{\mathrm{virt}}$ degrees of freedom
        that actually affect the energy.
        The RMS normalization makes the loss comparable across molecules of different size.

\section{Surrogate Metrics for Initial Guess Quality}
    \label{app:surrogate-metrics}

    Throughout this section, $\hat{\mathbf{X}}$ denotes a predicted (ML) quantity and $\mathbf{X}^*$ its converged SCF reference.
    ML models for SCF acceleration are typically trained and evaluated using elementwise error norms on the predicted tensors $\hat{\mathbf{P}}$ or $\hat{\mathbf{F}}$.
    These Frobenius-based targets are convenient to compute but do not directly characterize what makes an initial guess yield fast and stable SCF convergence.
    Here, we catalog a broader set of surrogate metrics that probe different aspects of initial-guess quality.

    \paragraph{Total energy.}
    The deviation in total electronic energy evaluated at the predicted density:
    \begin{align}
      \Delta E = |E(\hat{\mathbf{P}}) - E(\mathbf{P}^*)|, \qquad E(\mathbf{P}) = \sum_{\mu,\nu=1}^{B} H^{(\textrm{core})}_{\mu\nu}\,P_{\nu\mu} + \tfrac{1}{2}\sum_{\mu,\nu=1}^{B} J_{\mu\nu}(\mathbf{P})\,P_{\nu\mu} + E_{\mathrm{xc}}[\rho(\mathbf{P})],
    \end{align}
    where $H^{(\textrm{core})}_{\mu\nu}$ is the one-electron (core) Hamiltonian, $J_{\mu\nu}(\mathbf{P}) = \sum_{\lambda,\sigma=1}^{B}(\mu\nu|\lambda\sigma)\,P_{\lambda\sigma}$ is the Coulomb matrix, and $E_{\mathrm{xc}}$ is the exchange-correlation functional including optional HF-exchange contributions\footnote{Hence, $E_\textrm{xc}$ is formally a \emph{functional} of $\rho$ and, if the functional is a hybrid, also a \emph{function} of $\mathbf{P}$.}.
    Since $E(\mathbf{P})$ is stationary at $\mathbf{P}^*$, $\Delta E$ is second order in $\|\hat{\mathbf{P}} - \mathbf{P}^*\|$ and therefore insensitive to the direction of the initial guess error.

    \paragraph{Mean-field energy.}
    The sum of one-electron and classical Coulomb contributions,
    \begin{align}
      E_{\mathrm{MF}}(\mathbf{P}) = \sum_{\mu,\nu=1}^{B} H^{(\textrm{core})}_{\mu\nu}\,P_{\mu\nu} + \tfrac{1}{2}\sum_{\mu,\nu=1}^{B} J_{\mu\nu}(\mathbf{P})\,P_{\mu\nu},
    \end{align}
    provides an energetically meaningful measure of density quality that is independent of the choice of exchange-correlation functional~\citep{gouldStepDensityBenchmarking2023}.

    \paragraph{Real-space density norms.}
    The $L^p$ distance between densities:
    \begin{align}
      \|\Delta\rho\|_p = \Bigl(\int |\hat{\rho}(\mathbf{r}) - \rho^*(\mathbf{r})|^p\, d\mathbf{r}\Bigr)^{1/p}, \qquad p \in \{1, 2\}.
    \end{align}
    The $L^1$ norm gives the total absolute electron displacement; the $L^2$ norm, expressed in the AO basis, involves the four-center overlap tensor $(\mu\nu|\lambda\sigma) = \int \chi_\mu(\mathbf{r})\chi_\nu(\mathbf{r})\chi_\lambda(\mathbf{r})\chi_\sigma(\mathbf{r})\,d\mathbf{r}$.
    Both weight all spatial regions uniformly and do not distinguish OV errors from redundant OO or VV contributions.

    \paragraph{Dipole moment.}
    Unlike the $L^p$ norms, which weight all regions of space equally, the electronic dipole moment weights the density error by the position vector $\mathbf{r}$, making it a physically transparent measure of charge displacement:
    \begin{align}
      \Delta\mu = \|\boldsymbol{\mu}(\hat{\rho}) - \boldsymbol{\mu}(\rho^*)\|, \qquad \mu_\alpha(\rho) = -\int r_\alpha\, \rho(\mathbf{r})\, d\mathbf{r},
    \end{align}
    where the nuclear contribution cancels in the difference $\Delta\mu$ and is therefore omitted.
    The linear weighting by $\mathbf{r}$ projects out only the $\ell=1$ multipole component of $\Delta\rho$, so $\Delta\mu$ is insensitive to higher-order error structure, but directly governs long-range electrostatic properties.

    \paragraph{Orbital projection.}
    \citet{lehtolaAssessmentInitialGuesses2019} propose the projection of the initial guess onto the converged occupied subspace,
    \begin{align}
      Q = \sum_{\mu,\nu,\lambda,\kappa=1}^{B} \hat{P}_{\mu\nu}\,S_{\nu\lambda}\,P^*_{\lambda\kappa}\,S_{\kappa\mu},
    \end{align}
    as a continuous metric that separates initial guess quality from the dynamics of the SCF algorithm.

    \paragraph{DIIS residual.}
    The commutator that measures departure from self-consistency:
    \begin{align}
      r_{\mathrm{DIIS}} = \|\mathbf{R}\|_F, \qquad R_{\mu\nu} = \sum_{\lambda,\sigma=1}^{B}\Bigl(F_{\mu\lambda}(\hat{\mathbf{P}})\,\hat{P}_{\lambda\sigma}\,S_{\sigma\nu} - S_{\mu\lambda}\,\hat{P}_{\lambda\sigma}\,F_{\sigma\nu}(\hat{\mathbf{P}})\Bigr).
    \end{align}
    In an orthonormal basis this reduces to $\|[\mathbf{F}(\hat{\mathbf{P}}), \hat{\mathbf{P}}]\|_F$, whose OV block is the orbital gradient $g_{ia} = 2 F_{ia}$.
    This isolates the OV sector but treats all OV pairs uniformly regardless of their gap.

    \paragraph{Orbital rotation gradient.}
    The squared norm of the energy gradient with respect to occupied-virtual rotations:
    \begin{align}
      \|G\|^2 = \sum_{ia}\Bigl(\sum_{\mu,\nu=1}^{B} C^{\mathrm{occ}}_{\mu i}\,F_{\mu\nu}(\hat{\mathbf{P}})\,C^{\mathrm{virt}}_{\nu a}\Bigr)^{\!2},
    \end{align}
    where $\mathbf{C}^{\mathrm{occ}}$ and $\mathbf{C}^{\mathrm{virt}}$ are the occupied and virtual orbital coefficient matrices obtained from $\hat{\mathbf{P}}$.
    Like $r_{\mathrm{DIIS}}$, $\|G\|$ isolates the OV sector but weights all OV pairs uniformly regardless of their gap.

    \paragraph{Frobenius norms.}
    Elementwise matrix distances:
    \begin{align}
      \|\Delta \mathbf{F}\|_F = \Bigl(\sum_{\mu,\nu=1}^{B}|\hat{F}_{\mu\nu} - F^*_{\mu\nu}|^2\Bigr)^{\!1/2}, \qquad \|\Delta \mathbf{P}\|_F = \Bigl(\sum_{\mu,\nu=1}^{B}|\hat{P}_{\mu\nu} - P^*_{\mu\nu}|^2\Bigr)^{\!1/2}.
    \end{align}
    The Frobenius norm treats OO, OV, and VV blocks identically.
    For $\mathbf{F}$, core-core diagonal elements can dominate over the small but physically critical OV elements that enter the energy gradient $G_{ia}$.
    For $\mathbf{P}$, the set of valid density matrices forms a Grassmannian of dimension $N_{\mathrm{occ}} \times N_{\mathrm{virt}}$, far lower than the $B^2$ degrees of freedom of the full basis representation.
    The Frobenius norm penalizes deviations in all $B^2$ components equally, including OO and VV blocks that correspond to redundant rotations affecting neither observables nor convergence.

\section{Coefficient-Based Models}
    \label{app:coeff-fock}

    The auxiliary-coefficient ansatz predicts a linear expansion of the electron
    density,
    $\hat{\rho}(\mathbf{r}) = \hat{c}_P\,\chi^{(\mathrm{aux})}_P(\mathbf{r})$,
    rather than an AO density matrix~\citep{songNeuralSCFNeuralNetwork2024,liuUniversallyTransferableAcceleration2025}. The Kohn-Sham Fock
    matrix decomposes as
    \begin{align}
    \mathbf{F} \;=\; \mathbf{H}^{(\mathrm{core})}
    \;+\; \mathbf{J}[\rho]
    \;+\; \mathbf{V}^{(\mathrm{local/grid})}_{\mathrm{xc}}[\rho,\nabla\rho,\tau]
    \;+\; \alpha\,\mathbf{K}[\mathbf{P}] ,
    \label{eq:fock-split-coeff}
    \end{align}
    where $\alpha\in[0,1]$ is the exact-exchange fraction,
    $\mathbf{V}^{(\mathrm{local/grid})}_{\mathrm{xc}}$ collects the semi-local XC contribution,
    and $\mathbf{K}$ is the Hartree-Fock exchange matrix.
    $\mathbf{H}^{(\mathrm{core})}$ is density-independent, and $\mathbf{J}$ is evaluated
    from $\hat{\rho}$ through the standard auxiliary-basis density-fitting
    route~\citep{vahtrasIntegralApproximationsLCAOSCF1993}. The remaining two
    terms require inputs that $\hat{\rho}$ alone does not determine, namely the
    kinetic-energy density $\tau$ for meta-GGA functionals and the density matrix
    $\mathbf{P}$ for any functional with $\alpha>0$. \citet{liuUniversallyTransferableAcceleration2025}
    address each with a substitution derivable from $\hat{\rho}$.

    \paragraph{Meta-GGA functionals.}
    The kinetic-energy density
    \begin{align}
    \tau(\mathbf{r}) \;=\; \tfrac{1}{2}\sum_i \nabla\psi_i(\mathbf{r})\cdot\nabla\psi_i(\mathbf{r})
    \end{align}
    is a sum over occupied molecular orbitals $\{\psi_i\}$ and is not recoverable
    from $\rho$. \citet{liuUniversallyTransferableAcceleration2025} substitute the von Weizsäcker
    kinetic-energy density
    \begin{align}
    \tau_{\mathrm{vW}}(\mathbf{r}) \;=\; \frac{\nabla\rho(\mathbf{r})\cdot\nabla\rho(\mathbf{r})}{8\,\rho(\mathbf{r})} ,
    \label{eq:vw-tau}
    \end{align}
    which depends only on $\rho$ and $\nabla\rho$ and is therefore directly
    evaluable from $\hat{\mathbf{c}}$~\citep{perdewLaplacianlevelDensityFunctionals2007}. The $\tau$-dependent contribution to the
    meta-GGA XC matrix is then evaluated on a real-space quadrature grid
    $\{\mathbf{r}_g\}$ with weights $\{\omega_g\}$ as
    \begin{align}
    \bigl[\mathbf{V}^{(\tau)}_{\mathrm{xc}}\bigr]_{\mu\nu}
    =
    \frac{1}{2}\sum_g
    \omega_g\,
    v_\tau(\mathbf{r}_g)\,
    \nabla\chi_\mu(\mathbf{r}_g)\cdot\nabla\chi_\nu(\mathbf{r}_g),
    \qquad
    v_\tau
    =
    \frac{\partial [\rho\,\varepsilon_{\mathrm{xc}}(\rho,\nabla\rho,\tau)]}
    {\partial \tau}
    \bigg|_{\rho=\hat{\rho},\,\tau=\tau_{\mathrm{vW}}}.
    \end{align}
    $\tau_{\mathrm{vW}}$ is exact for single-orbital systems and introduces a
    systematic approximation error for general
    molecules~\citep{perdewLaplacianlevelDensityFunctionals2007}. The predicted
    $\hat{\rho}$ therefore parametrizes an approximate meta-GGA Fock build rather
    than the exact one.

    \paragraph{Hybrid and range-separated functionals.}
    The exchange matrix $\mathbf{K}[\mathbf{P}]$ is a function of the density matrix
    rather than of $\rho$. Many density matrices correspond to the same $\rho$,
    so reconstructing $\mathbf{P}$ from $\hat{\rho}$ is underdetermined.
    \citet{liuUniversallyTransferableAcceleration2025} replace the learned density matrix with the
    superposition-of-atomic-densities (SAD) matrix,
    \begin{align}
        \mathbf{P}_{\mathrm{SAD}} = \mathrm{blockdiag}(\mathbf{P}_{A_1}, \dots, \mathbf{P}_{A_M}) ,
        \label{eq:sad}
    \end{align}
    where $\mathbf{P}_A$ is the atomic SCF density matrix for atom $A$, and use
    $\mathbf{K}[\mathbf{P}_{\mathrm{SAD}}]$ in Eq.~\eqref{eq:fock-split-coeff}.
    $\mathbf{P}_{\mathrm{SAD}}$ coincides with the MINAO initial guess of most
    quantum-chemistry implementations~\citep{sunRecentDevelopmentsPySCF2020}. The $\alpha\,\mathbf{K}$ contribution is therefore independent of the learned prediction and coincides with the exchange contribution from a MINAO/SAD initial density, while the learned coefficients still determine the Coulomb and semi-local XC terms.

    \paragraph{ERIC considerations for coefficient models.}
    ERIC charges $\hat{\mathbf{c}} \to \mathbf{F}^{(-1)}$ as one full Fock build, matching the
    convention for matrix-based $\Delta$-learning. This is always an upper bound:
    $\mathcal{M}_{\mathbf{c}\to\mathbf{F}}$ skips the forward auxiliary fit $\mathbf{P}\to\mathbf{c}$
    that a normal $\mathbf{J}$ build requires, and substitutes $\tau_{\mathrm{vW}}$ for the MO-dependent $\tau$ in meta-GGAs at lower grid cost, so $\mathcal{M}_{\mathbf{c}\to\mathbf{F}}$ is
    strictly cheaper than $\mathcal{M}_{\mathbf{P}\to\mathbf{F}}$ in any reasonable implementation.
    The size of the gap depends on which terms dominate per-iteration cost.
    For semi-local functionals (PBE, SCAN), the iteration is dominated by $\mathbf{J}[\mathbf{P}]$
    at $\mathcal{O}(B^4)$ without density fitting and falls to $\mathcal{O}(B^2 B_{\mathrm{aux}})$
    with $J$-fitting, comparable to the grid-based $\mathbf{V}_{\mathrm{xc}}$ contribution; the
    $+1$ charge therefore overcounts substantially without DF and only modestly with DF.
    For hybrids, $\mathbf{J}$ and $\mathbf{K}$ each scale as $\mathcal{O}(B^4)$ without DF and
    contribute comparable wall-time, so the asymptotic initial-guess overhead reduces to roughly
    half a Fock build because it saves the $\mathbf{J}$ build of a standard iteration. With JK-fitting,
    $\mathbf{K}$ retains the larger prefactor due to its $N_{\mathrm{occ}}$-scaling and dominates
    the iteration cost; the coefficient overhead approaches (but stays below) a full Fock build.
    Across all six functional/DF combinations, the ERIC values we report for coefficient-based
    models are therefore conservative upper bounds, with the slack largest for semi-local
    functionals without DF and the smallest for hybrids with JK-fitting.

    \paragraph{MINAO overhead.}
    A separate wall-time consideration, independent of ERIC, is the cost of the MINAO (SAD) build itself.
    Matrix-based $\Delta$-learning requires it at every level of theory to construct the residual
    base $\mathbf{P}_{\mathrm{base}}$ for density-matrix targets, or
    $\mathbf{F}_{\mathrm{base}} = \mathcal{M}_{\mathbf{P}\to\mathbf{F}}(\mathbf{P}_{\mathrm{MINAO}})$
    for Hamiltonian targets. The coefficient model only references a density matrix through
    $\mathbf{K}[\mathbf{P}_{\mathrm{SAD}}]$ (Eq.~\eqref{eq:fock-split-coeff}), which drops out at
    $\alpha = 0$. At the GGA and meta-GGA level the coefficient pipeline therefore avoids the
    MINAO build entirely. At the hybrid level both pipelines pay it: $\Delta$-learning as the
    residual base, the coefficient model as $\mathbf{P}_{\mathrm{SAD}}$ for the exchange
    substitution.

\section{Evaluation Details}
    \label{app:evaluation details}
     The reference computations and final speedup evaluations are performed using PySCF~\citep{sunRecentDevelopmentsPySCF2020} and GPU4PySCF~\citep{liIntroducingGPUAcceleration2025} with grid level $1$. We use the def2-SVP basis set~\citep{weigendBalancedBasisSets2005} and do exact calculations.
     The RIC and ERIC are measured relative to PySCF's default (MINAO), using its default convergence threshold of $10^{-9} \:\textrm{Ha}$~\citep{sunRecentDevelopmentsPySCF2020} and maximum number of cycles $T=50$.

    \subsection{Data Split}
        We follow~\citet{yuQH9QuantumHamiltonian2024} and split QM9 by molecular size, assigning molecules with at most 20 atoms to training, 21--22 to validation, and 23 or more to testing. To evaluate out-of-distribution generalization beyond QM9, we additionally use QM40 and QMugs as far-out-of-distribution test sets. For QM40 we randomly sample up to 100 molecules per heavy-atom count, whereas for QMugs we sample 10 molecules at every tenth heavy-atom count from 40 to 90, drawing at most one conformer per molecule since QMugs provides three conformers each. For both QM40 and QMugs we restrict the sampling to molecules composed only of the elements present in QM9 (H, C, N, O, F), isolating size extrapolation from element extrapolation.

    \subsection{Compute Resources}
        \label{app:compute resources}
        All experiments were conducted on a local compute cluster. Pretraining and fine-tuning were performed on single-GPU jobs using NVIDIA H200 GPUs (144\,GB GPU memory). Pretraining runs additionally used 16 CPU cores and approximately 100\,GB of host RAM, while \ours{} fine-tuning used approximately 300\,GB of host RAM. Evaluation runs were conducted separately on a single NVIDIA A100 GPU with 40\,GB of memory.

        A single pretraining run required roughly 2.5--4 days of wall-clock time on one H200 GPU. In our setup, pretraining used less GPU memory than the default JAX preallocation, which is 75\% of total GPU memory. In practice, pretraining could also be carried out on A100-class GPUs without difficulty. A single SAIL fine-tuning run required roughly 1.5--3 days of wall-clock time on one H200 GPU and used approximately 87\% of the available H200 memory. Fine-tuning therefore required larger-memory GPUs than pretraining. Under the truncation ablations described in the paper, the fine-tuning memory requirement can be reduced.

        To reproduce the full set of reported results, we estimate a total budget of approximately
        $
        16 \times 3.5 + 22 \times 2.5 = 111
        $
        H200 GPU-days for pretraining and fine-tuning runs, plus approximately 10 A100 GPU-days for evaluation.
        The full research project required more compute than the totals above, since it also included failed runs, debugging, preliminary experiments, early-stage ablations, and hyperparameter exploration.

    \subsection{Wall-time measurement procedures}
    \label{app:wall-time measurement procedures}
    All wall-time measurements were performed on a single NVIDIA A100 GPU. We separate the neural initialization from the SCF loop into two stages rather than running them concurrently, because JAX and GPU4PySCF do not share a single GPU cleanly. JAX assumes it owns the device, tends to pre-allocate aggressively, and does not always release memory back to the driver, which interferes with GPU4PySCF's memory management. Running the two stages back-to-back on the same GPU, gives a clean measurement of each.

    In the first stage, the ML model produces an initial density matrix for each molecule. The reported initialization time covers the model forward pass and the construction of the initial density matrix from the model output. For the coefficient-based Fock build, we use the optimized implementation of~\citet{liuUniversallyTransferableAcceleration2025}. In the second stage, this density matrix is passed to GPU4PySCF for the SCF calculation using the same basis set, exchange-correlation functional, grid level, and density-fitting settings as the baseline. The accelerated wall-time is the sum of the two stages. The baseline is GPU4PySCF initialized with the standard MINAO guess.

    We exclude two categories of cost from the ML-initial guess acquisition, both of which a production implementation would avoid. First, we exclude the construction of integral tensors that GPU4PySCF recomputes in the subsequent SCF stage, such as the overlap matrix. A merged implementation would compute these once and share them across both stages. Second, we exclude JAX tracing and compilation, which ahead-of-time compilation removes entirely.

\section{Traditional Baselines}
    \begin{figure}[b]
        \centering
        \includegraphics[width=\textwidth]{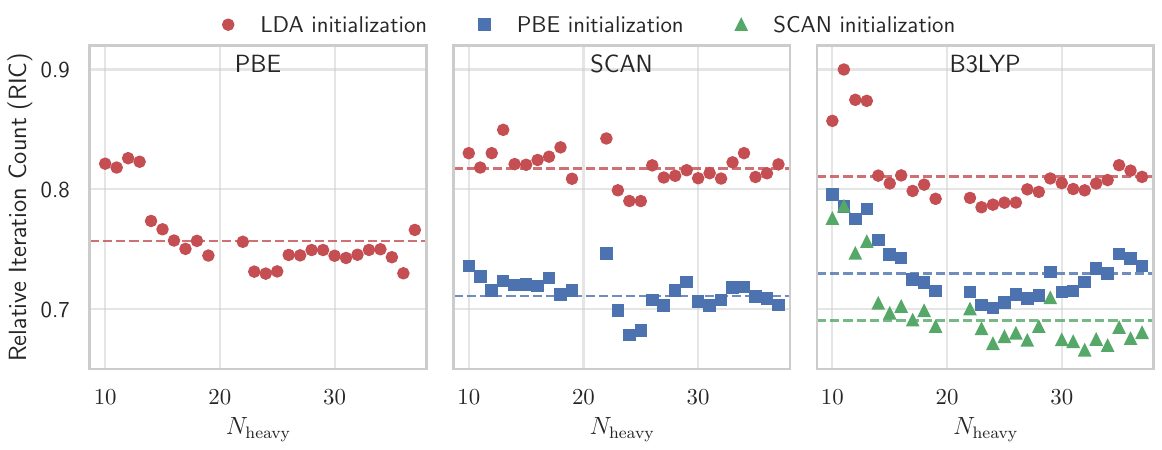}
        \caption{RIC when initializing KS-DFT from converged orbitals of a cheaper functional. Each panel is a different target functional, with initialization sources colored. Dashed lines show mean RIC. SCAN $\to$ B3LYP reaches the lowest RIC at ${\sim}70\%$.}
        \label{fig:traditional_baselines}
    \end{figure}
    \label{app:traditional baselines}
    A perfect prediction of the target ground state would reach RIC$=0$, so there is no upper bound on what a ground-state-based initial guess can achieve. What we can measure instead is how far a realistic but imperfect ground-state-like guess takes us. A natural way to construct such a guess is to run the target SCF starting from a converged ground-state density of a cheaper functional, where the imperfection comes from the difference between the two theories rather than from an ML approximation. This gives a reference point for the acceleration achievable from ground-state-like guesses that fall short of the true target by a known amount. We run all valid chains across Jacob's ladder (LDA $\to$ PBE, LDA/PBE $\to$ SCAN, and LDA/PBE/SCAN $\to$ B3LYP) and report the resulting RIC in Figure~\ref{fig:traditional_baselines}.

    Initializing from a converged cheaper functional reduces iteration counts by 20--30\% relative to MINAO, with higher-rung sources yielding progressively lower RIC. The reductions are stable across $N_\text{heavy}$. SCAN $\to$ B3LYP reaches the lowest RIC at ${\sim}70\%$.

    \textbf{Wall-time reductions do not follow.} The pre-run is not free. Even SCAN, despite its formally lower scaling than B3LYP, costs roughly ten Fock builds to converge, which must be offset by iteration savings in the subsequent hybrid SCF. An ML initial guess, by contrast, costs a single forward pass. The RIC reductions reported here are therefore not wall-time competitive with ML initialization, and serve only as a reference point for the iteration savings attainable when a ground-state-like guess approximates the target through a cheaper theory.

\section{Related Work and Additional Reference Models}
\label{app:prior-work-reference-RIC}

     \begin{table}[!b]
        \centering
        \caption{Relative iteration reduction ($1 - \textrm{RIC}$) with respect to MINAO as reported in prior work. Values with a star use $\Delta$-Learning for Fock prediction or a coefficient Fock build with similar cost, such that $\textrm{RIC} \neq \textrm{ERIC}$. OOD values in brackets are limited extrapolation.}
        \label{tab:prior-work-RIC}
        \small
        \renewcommand{\arraystretch}{1.25}
        \begin{tabular}{lccccp{4.4cm}}
        \toprule
        \textbf{Model} & \textbf{Theory} & \textbf{ID} & \textbf{OOD-S} & \textbf{OOD-L} & \textbf{Notes} \\
        \midrule
        \textit{QHNet} \\
        \citet{yuQH9QuantumHamiltonian2024} & B3LYP/SVP & $29\%$ & $28\%$ & - & QH9 dataset. For OOD: train and validation on $<22$, test on up to $29$ atoms \\
        \citet{zhangSelfConsistencyTrainingDensityFunctionalTheory2024} & " & - & $36\%$ & - & QH9 dataset, with subsampled test set. With additional self-consistency tuning \\
        \citet{zhangSelfConsistencyTrainingDensityFunctionalTheory2024} & " & $34\%^*$  & - & - & Trained on MD22 and QH9 datasets, and evaluated on MD22. With additional self-consistency tuning \\
        \citet{kimMachineLearningHamiltonians2026} & " & $33\%$ & - & - & QH9 dataset. Convergence tol. $10^{-13} \, \textrm{Ha}$ \\
        \citet{liEnhancingScalabilityApplicability2025} & "/TZVP & $10\%$ & - & - & PubChemQH (40 - 100 Atoms). WALoss \\
        \citet{liuUniversallyTransferableAcceleration2025} & PBE/SVP & $37\%$ & - & $-80\%$ & On SCF-bench (QM9-like) \\
        \midrule
        \textit{WANet} \\
        \citet{liEnhancingScalabilityApplicability2025} & B3LYP/TZVP & $18\%$ & - & - & PubChemQH (40 - 100 Atoms). WALoss \\
        \midrule
        \textit{QHFlow} \\
        \citet{kimMachineLearningHamiltonians2026} & B3LYP/SVP & $40\%^*$ & - & - & QH9 dataset. Convergence tol. $10^{-13} \, \textrm{Ha}$ \\
        \citet{liuUniversallyTransferableAcceleration2025} & PBE/SVP & $43\%^*$ & - & $-47\%^*$ & On SCF-bench (QM9-like) \\
        \midrule
        \textit{Graph2Mat} \\
        \citet{febrerGraph2MatUniversalGraph2025} & PBE/DZP & $40\%$ & - & - & On QM9 molecules with SIESTA defaults~\citep{garciaSiestaRecentDevelopments2020}. PBE functional confirmed by correspondence with the authors.\\
        \midrule
        \textit{aux-coefficients} \\
        \citet{liuUniversallyTransferableAcceleration2025} & PBE/SVP & $36\%^*$ &-& $33\%^*$ & On SCF-bench (QM9-like) \\
        " & SCAN/ " & $12\%^*$ &-& $14\%^*$ & " trained on PBE \\
        " & B3LYP/ " & $15\%^*$ &-& $16\%^*$ & " trained on PBE \\
        \bottomrule
        \end{tabular}
    \end{table}

    \paragraph{OOD-S vs OOD-L.}
    Here we introduce two out-of-distribution regimes to reconcile our results with prior work.
    OOD-S follows the standard QH9-stable-ood split of~\citet{yuQH9QuantumHamiltonian2024}, in which training molecules have up to 20 atoms and test molecules contain up to 29 atoms, corresponding to a mild size extrapolation of less than $1.5\times$ the training distribution.
    OOD-L corresponds to a more demanding protocol in which a substantial fraction of the test set consists of molecules more than twice the size of the largest training molecule.
    Since the OOD split of QH9 alone has been reported to be insufficient to expose size-transferability failures of learned Hamiltonian initializations~\citep{liuUniversallyTransferableAcceleration2025}, we adopt OOD-L as our primary out-of-distribution benchmark and retain OOD-S only for comparability with prior work.
    We match the metric used in prior work (RIC; Table~\ref{tab:prior-work-RIC}) in our own measurements (Table~\ref{tab:this-work-RIC}), expressing both as the relative iteration reduction $1 - \textrm{RIC}$ with respect to MINAO so that readers can compare across the two tables directly.

\begin{table}[t]
    \centering
    \caption{Relative iteration reduction ($1 - \textrm{RIC}$) with respect to MINAO, measured in this work. Higher is better; negative values indicate deceleration. We use the def2-SVP basis set and PySCF's defaults~\citep{sunRecentDevelopmentsPySCF2020}. The Coefficients, Density Matrix, and Fock rows are trained on QM9 with OOD-L evaluated on QM40. Values with a star use $\Delta$-Learning for Fock prediction or a coefficient Fock build with similar cost, such that $\textrm{RIC} \neq \textrm{ERIC}$.}
    \label{tab:this-work-RIC}
    \small
    \renewcommand{\arraystretch}{1.25}
    \begin{NiceTabular}{llccccc}
    \CodeBefore
    \rectanglecolor{sailblue}{2-4}{\arabic{iRow}-4}
    \rectanglecolor{sailblue}{2-7}{\arabic{iRow}-7}
    \Body
    \toprule
    \textbf{Model} & \textbf{Functional} & \multicolumn{2}{c}{\textbf{OOD-S}} & & \multicolumn{2}{c}{\textbf{OOD-L}} \\
    \cmidrule(lr){3-4} \cmidrule(lr){6-7}
    & & base & SAIL & & base & SAIL \\
    \midrule
    \textit{$\Delta$-QHNet} \\
    \textbf{This work} & B3LYP & $30\%^*$ & - & & $-94\%^*$ & - \\
    \midrule
    \textit{QHFlow} \\
    \textbf{This work} & B3LYP & $32\%^*$ & - & & $-88\%^*$ & - \\
    \midrule
    \textit{QDensFlow} \\
    \textbf{This work} & B3LYP & $15\%$ & - & & $-22\%$ & - \\
    \midrule
    \midrule
    \textit{Coefficients} & PBE   & $44\%^*$ & $41\%^*$ & & $42\%^*$ & $41\%^*$ \\
                    & SCAN  & $18\%^*$ & $31\%^*$ & & $23\%^*$ & $33\%^*$ \\
                    & B3LYP & $13\%^*$ & $33\%^*$ & & $18\%^*$ & $32\%^*$ \\
    \midrule
    \multirow{3}{*}{\textit{Density Matrix}}
                    & PBE   & $29\%$ & $29\%$ & & $-5\%$ & $40\%$ \\
                    & SCAN  & $29\%$ & $36\%$ & & $1\%$  & $35\%$ \\
                    & B3LYP & $25\%$ & $35\%$ & & $-2\%$ & $32\%$ \\
    \midrule
    \multirow{3}{*}{\textit{Fock}}
                    & PBE   & $32\%^*$ & $28\%^*$ & & $-11\%^*$ & $34\%^*$ \\
                    & SCAN  & $34\%^*$ & $33\%^*$ & & $-16\%^*$ & $31\%^*$ \\
                    & B3LYP & $32\%^*$ & $29\%^*$ & & $-47\%^*$ & $30\%^*$ \\
    \bottomrule
    \end{NiceTabular}
\end{table}

    \paragraph{Reproduction of QHFlow.}
    Using the official QHFlow reference implementation~\citep{kimHighorderEquivariantFlow2025}, we are unable to reproduce the $69\%$ ID SCF iteration reduction originally reported, and our numbers instead align with the independent evaluation of~\citet{liuUniversallyTransferableAcceleration2025} and with the authors' own re-evaluation in~\citet{kimMachineLearningHamiltonians2026}, both of which report substantially lower acceleration.
    Inspection of the reference implementation reveals two accounting choices that inflate the reported acceleration relative to a like-for-like comparison against the MINAO baseline used elsewhere in the literature~\citep{sunRecentDevelopmentsPySCF2020}.
    First, although the paper describes the $100\%$ reference as conventional DFT initialized with MINAO, the inference pipeline computes the baseline SCF using the \texttt{1e} (core-Hamiltonian) guess via \texttt{init\_guess\_by\_1e},\footnote{\url{https://github.com/seongsukim-ml/QHFlow/blob/b2e8662ee8e6549e2a00b66bb41f0709bf24a4b3/src/pl_module/base_module.py\#L356}, commit \texttt{b2e8662}, dated 2025-10-25.} which is known to be a strictly worse starting point than MINAO~\citep{lehtolaAssessmentInitialGuesses2019} and therefore inflates the denominator of every iteration and wall-time ratio.
    Second, QHFlow is a residual-learning model and requires an initial Fock matrix $F_\textrm{init}$ as input; this $F_\textrm{init}$ is produced by a full Fock build on top of a MINAO density\footnote{\url{https://github.com/seongsukim-ml/QHFlow/blob/b2e8662ee8e6549e2a00b66bb41f0709bf24a4b3/src/dataset_module/ori_dataset.py\#L313-L330}, same commit.} and cached in the preprocessed dataset, so that the per-molecule cost of constructing the model's own input is amortized into offline dataset preparation and is absent from both the inference timer and the SCF timer.
    A similar gap applies to the generalized eigendecomposition that converts the predicted Hamiltonian into an initial density matrix, which is executed after the inference timer has stopped and before the SCF timer has started.
    Under an evaluation protocol that charges these costs to the per-molecule budget and uses MINAO as the $100\%$ reference, we obtain the numbers reported in Table~\ref{tab:this-work-RIC}.
    We can, however, reproduce the size extrapolation failure mode reported by~\citet{liuUniversallyTransferableAcceleration2025} for the original QHFlow.

    \textbf{QDensFlow.}
    To disentangle the cost of the extra Fock build required by Hamiltonian-target learning from the acceleration attributable to the learned initialization itself, we introduce QDensFlow, an adapted variant of QHFlow in which the supervision target is swapped from the Fock matrix to the density matrix while keeping the architecture, prior distributions, flow-matching objective, and residual-learning setup otherwise unchanged.
    The predicted density matrix is passed directly to SCF as \texttt{dm0}, and the MINAO density matrix used as the residual reference $P_\textrm{init}$ is available at negligible cost, removing the per-molecule Fock build that Hamiltonian-target residual learning requires.
    We do not expect this swap to resolve the size extrapolation failure and confirm this empirically in Table~\ref{tab:this-work-RIC}, but it allows a fair accounting of the per-molecule cost of learned initialization within the QHFlow framework.

    \paragraph{Consistency with independent evaluations.}
    Our reproduction is consistent with the independent evaluation of~\citet{liuUniversallyTransferableAcceleration2025}, who report a QHFlow ID acceleration of $43\%$ and an OOD-L deceleration of $-47\%$ on SCF-bench, well below the $69\%$/$68\%$ originally claimed.
    The authors of QHFlow themselves appear to have revised these numbers downward in their follow-up work~\citep{kimMachineLearningHamiltonians2026}, where QHFlow is re-evaluated under a shared SCF protocol and reaches only $40\%$ ID acceleration on QH9 (B3LYP/SVP), in line with our measurements and with those of~\citet{liuUniversallyTransferableAcceleration2025}.

\newpage
\section{Convergence Thresholds}
    \label{app:convergence}

\begin{figure}[h]
      \centering
      \vspace{-0.2cm}
      \includegraphics[width=0.8\textwidth]{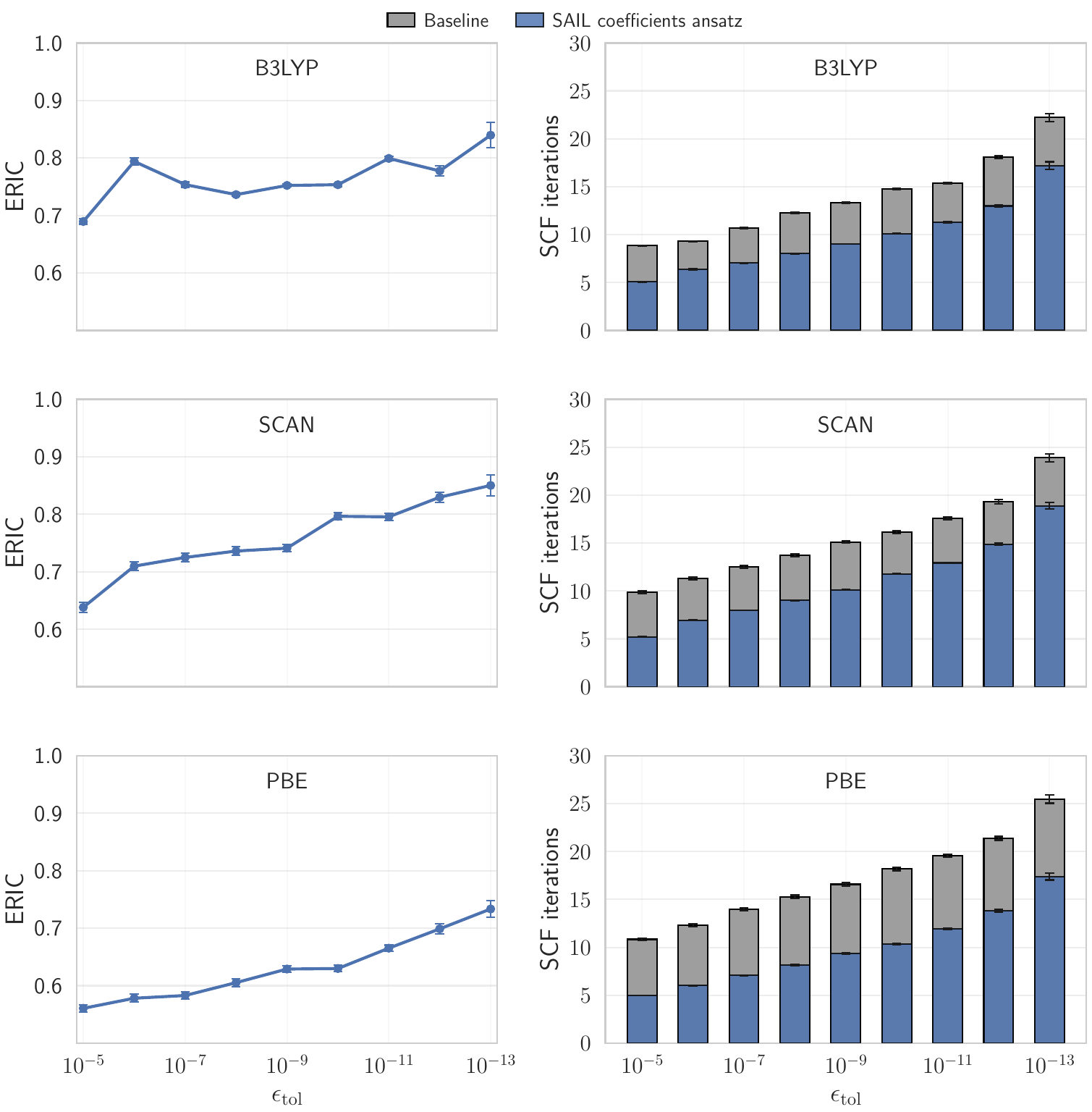}
      \vspace{-0.2cm}
      \caption{%
      Effect of the SCF convergence threshold $\epsilon_\text{tol}$ on accuracy and
      computational cost, evaluated on QM40 molecules with heavy atom count 36.
      \textbf{Left column:} ERIC as a function of $\epsilon_\text{tol}$ for B3LYP (top),
      SCAN (middle), and PBE (bottom). Tighter thresholds yield higher ERIC across all
      functionals, with the variation between 0.15 and 0.21 ERIC units over the full range.
      \textbf{Right column:} Mean number of SCF iterations, decomposed into the baseline
      contribution (gray) and the overhead of the SAIL coefficient ansatz (blue).
      Iteration counts roughly double from $\epsilon_\text{tol} = 10^{-5}$ to $10^{-13}$.
      Error bars indicate the standard error over the evaluation set.
      All main-paper results use $\epsilon_\text{tol} = 10^{-9}$ (vertical dashed line
      in the main text figures).
    }
    \label{fig:convergence}
    \end{figure}

    The SCF convergence threshold $\epsilon_\text{tol}$ governs when the self-consistent field
    iteration is considered converged, and it has a non-trivial effect on both the quality
    of the resulting electron density and the computational cost incurred.
    Figure~\ref{fig:convergence} shows this dependence for B3LYP, SCAN, and PBE on the QM40
    benchmark at heavy atom count 36, sweeping $\epsilon_\text{tol}$ from $10^{-5}$ to
    $10^{-13}$.

    \paragraph{Effect on SCF cost.}
    The right column of Figure~\ref{fig:convergence} decomposes the mean number of SCF
    iterations into the baseline contribution and the number of iterations the \ours{}
    coefficient ansatz requires. Both components grow with tighter thresholds, and the total iteration
    count roughly doubles when moving from $\epsilon_\text{tol} = 10^{-5}$ to
    $10^{-13}$ (e.g., from $\approx 9$ to $\approx 22$ for B3LYP).

    \looseness=-3
    \textbf{Choice of threshold and cross-work comparisons.} Throughout the main paper we use $\epsilon_\text{tol} = 10^{-9}$, which offers a favorable balance between converged densities and computational cost. We stress that comparisons across different works must account for the convergence threshold used: a method evaluated at $\epsilon_\text{tol} = 10^{-6}$ will systematically report lower ERIC values than the same method at $10^{-10}$, regardless of the underlying functional quality. When reproducing or comparing to our results, we therefore recommend using the same threshold, or reporting results at multiple thresholds as shown here.

\newpage
\section{Hyperparameters}
    \label{app:hyperparameters}
    \begin{table}[H]
    \centering
    \small
    \setlength{\tabcolsep}{5pt}
    \renewcommand{\arraystretch}{1.15}
    \caption{Hyperparameters not listed here follow the defaults of EquiformerV2 (base model), NeuralSCF (embedding and coefficient readout), and QHNet (matrix readout).}
    \begin{tabular}{p{0.28\linewidth} >{\raggedleft\arraybackslash}p{0.16\linewidth} p{0.52\linewidth}}
        \toprule
        \textbf{Parameter} & \textbf{Value} ($\mathbf{c}$/ $\mathbf{P}$/ $\mathbf{F}$) & \textbf{Notes} \\
        \midrule
        \multicolumn{3}{l}{\textit{Base Model}} \vspace{0.1cm} \\
        Radial cutoff (\AA)     & $5$ & Typical value for MLIPs, also used by Equiformerv2.\\
        Emb dim encoder & $128$ & Dimensionality of each learned atom-type embedding. \\
        Env hidden encoder      & $64$ & Width of the encoder edge MLP. \\
        Gaussian RBF  & $128$ & \\
        Resolution of point samples $R$                          & 9        & We use $\beta = 9$, $\alpha = 2\beta +1$ as opposed to square 18x18. Memory-costly hyperparameter with negligible difference for higher values.\\
        Maximum degree $L_{max}$ & $4$ & Matches highest angular momentum of (auxiliary) basis functions used in our experiments \\
        \midrule
        \multicolumn{3}{l}{\textit{Baseline (Pretraining)}} \vspace{0.1cm}\\
        Loss        & $L^2$--Overlap/ mixed Frobenius--$L^1$ &  We used the proposed loss by~\citet{songNeuralSCFNeuralNetwork2024} for our coefficient-based models and~\citet{yuQH9QuantumHamiltonian2024} for the matrix-based models.\\
        Optimizer          & Muon      & More stable, less sensitive to LR~\citep{jordan2024muon}.\\
        Epochs & $100$ & Good balance between performance and training time. \\
        Batch size & $32$ & Tried \{$1,4,8,32,64$\}. \\
        Warmup schedule             & linear    & $1000$ Steps.\\
        Decay              & Cosine &  \\
        Base LR            & 5e-3       & Tried \{2e-2, 1e-2, 5e-3, 2e-3, 1e-3, 5e-4\}. LR higher than 5e-3 would sometimes result in divergence. \\
        Min LR             & 1e-6            &  \\
        EMA decay             & $0.995$        & Exponential moving average of model parameters. Stabilizes training, improves generalization. \\
        Gradient clipping norm   & $10.0$ & Reduces impact of outliers. \\
        Weight decay       & 1e-3       &  \\
        \midrule
        \multicolumn{3}{l}{\textit{\ours{} (Fine-tuning)}} \vspace{0.1cm} \\
        Loss        & Eq.~\eqref{eq:SAIL-loss} & Commutator Loss can be used as an alternative. \\
        Optimizer          & Muon           &  ~\citep{jordan2024muon}\\
        Epochs             & $5$              &  \\
        Batch size & $1$ & Current implementation does not support batching\\
        Warmup schedule            & linear         & $1000$ Steps.\\
        Decay              & Cosine         &  \\
        Base LR            & 5e-4 / 1e-3 / 5e-4      & Tried \{5e-3, 2e-3, 1e-3, 5e-4, 2e-4, 1e-4, 5e-5, 1e-5\}. While lower LR can have better learning curves ID, they end up performing worse OOD. We emphasize the importance of higher LR in combination with EMA and Muon.  \\
        Min LR             & 1e-7           & Improvements unnoticeable below this. \\
        EMA decay             & $0.995$        & Exponential moving average of model parameters. Stabilizes training, improves generalization. \\
        Gradient clipping norm   & $1.0$      & \\
        SCF-cycles $T$     &      $10$          &  Most molecules converge within $10$ cycles. Ablations showed that it can be reduced to $T=1$ without losing significant performance.\\
        SCF-loss weighting    &  Uniform  & Since early SCF iterations typically have larger errors, this naturally gives them a larger absolute contribution to the total loss. We also tested reweighting schemes that equalize relative contributions across cycles or emphasize later cycles, but both degraded performance. \\
        \bottomrule
    \end{tabular}
    \end{table}

\newpage
\section{Further Ablations}
    \label{app:ablations}
    \begin{table}[h]
        \centering
        \caption{
        \textbf{SCF acceleration of baseline and \ours{} (fine-tuned) models} across input representations and functionals, evaluated on QM40 (out-of-distribution). \textcolor{gray}{Grey rows} show ablation variants. \textbf{Bold}: best RIC/ERIC per functional across all input representations.}
        \label{tab:sail_full}
        \small
        \begin{NiceTabular}{lll l cc cc cc}
        \CodeBefore
        \rectanglecolor{sailblue}{2-6}{\arabic{iRow}-6}
        \rectanglecolor{sailblue}{2-8}{\arabic{iRow}-8}
        \rectanglecolor{sailblue}{2-10}{\arabic{iRow}-10}
        \Body
        \toprule
        \textbf{Model} & & \textbf{Functional} & \textbf{Ablation} & \multicolumn{2}{c}{\textbf{Surrogate Loss} ($\downarrow$)} & \multicolumn{2}{c}{\textbf{RIC} ($\downarrow$)} & \multicolumn{2}{c}{\textbf{ERIC} ($\downarrow$)} \\
        & & & & base & \ours{} & base & \ours{} & base & \ours{} \\
        \midrule
        Coefficients & $\mathbf{c}$ & PBE & & $0.003$ & $0.015$ & $0.58$ & $0.59$ & $0.65$ & $0.66$ \\
        & & \textcolor{gray}{"} & \textcolor{gray}{Embedding} & \textcolor{gray}{$0.007$} & \textcolor{gray}{$0.013$} & \textcolor{gray}{$0.62$} & \textcolor{gray}{$0.60$} & \textcolor{gray}{$0.69$} & \textcolor{gray}{$0.67$} \\
        & & \textcolor{gray}{"} & \textcolor{gray}{$B_{\mathrm{aux}}$} & \textcolor{gray}{$0.005$} & \textcolor{gray}{$0.014$} & \textcolor{gray}{$\mathbf{0.55}$} & \textcolor{gray}{$0.61$} & \textcolor{gray}{$0.62$} & \textcolor{gray}{$0.67$} \\
        & & SCAN & & $0.004$ & $17.558$ & $0.77$ & $0.67$ & $0.85$ & $0.74$ \\
        & & \textcolor{gray}{"} & \textcolor{gray}{$B_{\mathrm{aux}}$} & \textcolor{gray}{$0.005$} & \textcolor{gray}{$21.403$} & \textcolor{gray}{$0.76$} & \textcolor{gray}{$0.67$} & \textcolor{gray}{$0.84$} & \textcolor{gray}{$0.74$} \\
        & & B3LYP & & $0.003$ & $53.815$ & $0.82$ & $\mathbf{0.68}$ & $0.90$ & $0.76$ \\
        & & \textcolor{gray}{"} & \textcolor{gray}{$B_{\mathrm{aux}}$} & \textcolor{gray}{$0.003$} & \textcolor{gray}{$61.490$} & \textcolor{gray}{$0.82$} & \textcolor{gray}{$\mathbf{0.68}$} & \textcolor{gray}{$0.89$} & \textcolor{gray}{$0.76$} \\
        \midrule
        Density Mat. & $\mathbf{P}$ & PBE & & $0.281$ & $1.149$ & $1.05$ & $0.60$ & $1.05$ & $\mathbf{0.60}$ \\
        & & SCAN & & $0.250$ & $1.100$ & $0.99$ & $\mathbf{0.65}$ & $0.99$ & $\mathbf{0.65}$ \\
        & & \textcolor{gray}{"} & \textcolor{gray}{$\mathcal{L}_{\mathrm{comm}}$} & \textcolor{gray}{"} & \textcolor{gray}{$1.108$} & \textcolor{gray}{"} & \textcolor{gray}{$0.65$} & \textcolor{gray}{"} & \textcolor{gray}{$0.65$} \\
        & & \textcolor{gray}{"} & \textcolor{gray}{T=4} & \textcolor{gray}{"} & \textcolor{gray}{$1.101$} & \textcolor{gray}{"} & \textcolor{gray}{$0.65$} & \textcolor{gray}{"} & \textcolor{gray}{$0.65$} \\
        & & \textcolor{gray}{"} & \textcolor{gray}{T=2} & \textcolor{gray}{"} & \textcolor{gray}{$1.110$} & \textcolor{gray}{"} & \textcolor{gray}{$0.65$} & \textcolor{gray}{"} & \textcolor{gray}{$0.65$} \\
        & & \textcolor{gray}{"} & \textcolor{gray}{T=1} & \textcolor{gray}{"} & \textcolor{gray}{$1.145$} & \textcolor{gray}{"} & \textcolor{gray}{$0.65$} & \textcolor{gray}{"} & \textcolor{gray}{$0.65$} \\
        & & \textcolor{gray}{"} & \textcolor{gray}{Non-$\Delta$} & \textcolor{gray}{$0.260$} & \textcolor{gray}{$0.942$} & \textcolor{gray}{$1.09$} & \textcolor{gray}{$0.66$} & \textcolor{gray}{$1.09$} & \textcolor{gray}{$0.66$} \\
        & & \textcolor{gray}{"} & \textcolor{gray}{Single-stage} & \textcolor{gray}{$-$} & \textcolor{gray}{$1.270$} & \textcolor{gray}{$-$} & \textcolor{gray}{$0.66$} & \textcolor{gray}{$-$} & \textcolor{gray}{$0.66$} \\
        & & B3LYP & & $0.267$ & $0.976$ & $1.02$ & $\mathbf{0.68}$ & $1.02$ & $\mathbf{0.68}$ \\
        \midrule
        Fock & $\mathbf{F}$ & PBE & & $0.005$ & $0.277$ & $1.11$ & $0.66$ & $1.18$ & $0.73$ \\
        & & SCAN & & $0.017$ & $0.284$ & $1.16$ & $0.69$ & $1.23$ & $0.76$ \\
        & & \textcolor{gray}{"} & \textcolor{gray}{Non-$\Delta$} & \textcolor{gray}{$0.453$} & \textcolor{gray}{$3.431$} & \textcolor{gray}{$1.99$} & \textcolor{gray}{$0.79$} & \textcolor{gray}{$1.99$} & \textcolor{gray}{$0.79$} \\
        & & B3LYP & & $0.026$ & $0.382$ & $1.47$ & $0.70$ & $1.55$ & $0.78$ \\
        & & \textcolor{gray}{"} & \textcolor{gray}{Non-$\Delta$} & \textcolor{gray}{$0.185$} & \textcolor{gray}{$6.913$} & \textcolor{gray}{$2.08$} & \textcolor{gray}{$0.74$} & \textcolor{gray}{$2.08$} & \textcolor{gray}{$0.74$} \\
        \bottomrule
        \end{NiceTabular}
        \vspace{-0.1cm}
    \end{table}

    We ablate the density embedding of~\citet{songNeuralSCFNeuralNetwork2024}, which conditions the equivariant node features on the MINAO density.
    Removing it improves both the baseline and SAIL fine-tuned models.
    Replacing the default \texttt{jfit} with the larger \texttt{jkfit} auxiliary basis moves the \ours{}-trained ERIC by $\leq 0.02$ on every functional while substantially raising the cost of the initial Fock build $\mathcal{M}_{\mathbf{c} \to \mathbf{F}}$, so we keep the cheaper default.
    As an alternative to the gradient loss $\mathcal{L}_{\nabla}$ (Eq.~\ref{eq:SAIL-loss}), we have tried a commutator-based loss
    \begin{align}\label{eq:comm-loss}
        \mathcal{L}_{\mathrm{comm}}^{(t)}
          = \frac{1}{B}\,
            \left\|\,
              \mathbf{F}^{(t)}\,\mathbf{P}^{(t)}\,\mathbf{S}
              - \mathbf{S}\,\mathbf{P}^{(t)}\,\mathbf{F}^{(t)}
            \right\|_{F},
    \end{align}
    which corresponds to the root-mean-square of the DIIS error matrix (Appendix~\ref{app:diis}), a convergence measure routinely used in quantum chemistry solvers as the extrapolation residual~\citep{pulayConvergenceAccelerationIterative1980,pulayImprovedSCFConvergence1982}.
    Both losses yield identical ERIC ($0.65$) on SCAN. Energy-based alternatives, such as the energy change between consecutive iterations or the energy difference to the final iterate, failed to produce any acceleration in our early experiments.

    Across functionals, $\Delta$-density matrix prediction emerges as a strong default for SCF acceleration. On PBE, it achieves ERIC $0.60$ vs.\ $0.66$ for the coefficient approach while generalizing to all functional classes without modification.
    For SCAN and B3LYP, the coefficient baseline reaches ERICs of $0.85$ and $0.90$. \ours{} recovers this to $0.74$ and $0.76$, still behind $\Delta$-density matrix ($0.65$ and $0.68$).
    Although $\Delta$-Fock prediction achieves a low RIC on B3LYP ($0.70$), its hidden Fock build yields an ERIC of $0.78$, behind the density matrix model at $0.68$.

\newpage
\section{Generalization to Transition States}
    \label{app:transition-states}

    \begin{wraptable}{r}{0.6\textwidth}
        \centering
        \vspace{-0.7cm}
        \setlength{\tabcolsep}{4pt}
        \caption{%
            \textbf{Intramolecular H-transfer} in $\mathrm{C_3H_8N_2O}$ from Transition1x~\citep{schreinerTransition1xDatasetBuilding2022} at B3LYP/def2-SVP using the single-stage ($T=1$) SPICEv2 $\mathbf{P}$ model.
        }
        \label{tab:transition-states-h-transfer}
        \vspace{0.1cm}
        \begin{NiceTabular}{clccc}
            \CodeBefore
            \rectanglecolor{sailblue}{2-5}{\arabic{iRow}-5}
            \Body
            \toprule
            $\xi$ & \textbf{Frame} & $\Delta E$ & \textbf{MINAO} & \textbf{ERIC} ($\downarrow$) \\
            & & (kcal/mol) & (iter.) & \ours{} \\
            \midrule
            $0.00$ & reactant         & $0.00$  & $11$ & $0.64$ \\
            $0.33$ &                  & $2.05$  & $11$ & $0.73$ \\
            $\dots$ &                 & $\dots$ & $\dots$ & $\dots$ \\
            $0.43$ &                  & $6.25$  & $11$ & $0.64$ \\
            $0.45$ &                  & $13.48$ & $12$ & $0.67$ \\
            $0.46$ &                  & $24.30$ & $12$ & $0.67$ \\
            $0.48$ &                  & $48.82$ & $10$ & $0.90$ \\
            $0.49$ &                  & $48.84$ & $11$ & $0.82$ \\
            $0.50$ & transition state & $48.83$ & $11$ & $0.82$ \\
            $0.51$ &                  & $48.83$ & $10$ & $0.90$ \\
            $0.60$ &                  & $37.47$ & $11$ & $0.73$ \\
            $0.64$ &                  & $27.20$ & $10$ & $0.80$ \\
            $\dots$ &                 & $\dots$ & $\dots$ & $\dots$ \\
            $0.89$ &                  & $20.15$ & $10$ & $0.80$ \\
            $1.00$ & product          & $19.78$ & $10$ & $0.80$ \\
            \bottomrule
        \end{NiceTabular}
        \centering
        \setlength{\tabcolsep}{3.5pt}
        \caption{%
            \textbf{Bond dissociation} $\mathrm{C_7H_{16} \to C_6H_{12} + CH_4}$ from Transition1x~\citep{schreinerTransition1xDatasetBuilding2022} at B3LYP/def2-SVP using the single-stage ($T=1$) SPICEv2 $\mathbf{P}$ model.
        }
        \label{tab:transition-states-dissociation}
        \vspace{0.2cm}
        \small
        \begin{NiceTabular}{clcccc}
            \CodeBefore
            \rectanglecolor{sailblue}{2-6}{\arabic{iRow}-6}
            \Body
            \toprule
            $\xi$ & \textbf{Frame} & $\mathrm{C{-}C}$ & $\Delta E$ & \textbf{MINAO} & \textbf{ERIC} ($\downarrow$) \\
            & & (\AA) & (kcal/mol) & (iter.) & \ours{} \\
            \midrule
            $0.00$ & $\mathrm{C_7H_{16}}$         & $1.53$ & $0.00$   & $9$  & $0.67$ \\
            $0.07$ &                              & $1.55$ & $2.00$   & $9$  & $0.67$ \\
            $0.20$ &                              & $1.65$ & $25.96$  & $9$  & $0.78$ \\
            $0.33$ &                              & $2.03$ & $79.14$  & $10$ & $0.80$ \\
            $0.48$ &                              & $2.23$ & $103.15$ & $10$ & $0.90$ \\
            $0.50$ & transition state             & $2.27$ & $103.22$ & $10$ & $0.90$ \\
            $0.52$ &                              & $2.27$ & $103.29$ & $10$ & $0.90$ \\
            $0.60$ &                              & $2.42$ & $103.48$ & $10$ & $0.90$ \\
            $0.90$ &                              & $2.95$ & $95.39$  & $9$  & $1.00$ \\
            $1.00$ & $\mathrm{C_6H_{12} + CH_4}$  & $3.13$ & $94.59$  & $9$  & $1.00$ \\
            \bottomrule
        \end{NiceTabular}
        \vspace{-0.4cm}
    \end{wraptable}

    To test out-of-distribution generalization to reactive non-equilibrium structures, we evaluate our single-stage ($T=1$) density-matrix model trained on SPICEv2 at the B3LYP/def2-SVP level of theory on two reaction pathways from the Transition1x dataset~\citep{schreinerTransition1xDatasetBuilding2022}.
    Along each pathway, $\xi = 0$ denotes the reactant, $\xi = 0.5$ the labeled transition state, and $\xi = 1$ the product, parameterized by the signed RMSD to the transition state and normalized separately on each side.

    Table~\ref{tab:transition-states-h-transfer} reports results for intramolecular hydrogen transfer in $\mathrm{C_3H_8N_2O}$.
    \ours{} accelerates every frame along the reaction coordinate (ERIC $0.64$--$0.90$), with the margin narrowing near the transition point ($\Delta E \approx 48.8\,\mathrm{kcal/mol}$, ERIC $0.82$--$0.90$).

    Table~\ref{tab:transition-states-dissociation} evaluates the $\mathrm{C{-}C}$ bond dissociation of the largest molecule in Transition1x, $\mathrm{C_7H_{16} \to C_6H_{12} + CH_4}$.
    At the equilibrium bond length ($1.53\,$\AA), the structure is in distribution and reaches an ERIC of $0.67$.
    As the $\mathrm{C{-}C}$ bond stretches, ERIC increases to $0.90$ by $2.23\,$\AA{} around the transition state and reaches $1.00$ at $2.95$--$3.13\,$\AA{} once the bond is broken.
    Although SPICEv2 contains many multi-fragment systems, it includes no bonds caught mid-dissociation, and at $3.13\,$\AA{} the fragments remain within the $\mathrm{C\cdots C}$ van der Waals contact ($\approx 3.4\,$\AA) as a compressed contact pair rather than a separated dimer.
    Even in this regime, the \ours{} initial guess never decelerates the solver relative to MINAO.

\newpage
\section{Robustness of \ours{}}
    \label{app:robustness}

    \paragraph{Training-seed variability.}
    To assess variability across training seeds, we retrain the B3LYP coefficient and density-matrix models, both the surrogate-loss baselines and their SAIL fine-tunes, with three training seeds each and evaluate them on QM40 and QMugs (Table~\ref{tab:seeds}). Across SAIL-trained models, the seed-to-seed standard deviation of the mean ERIC lies between $0.001$ and $0.010$. SAIL also makes models more reproducible: the density-matrix baseline varies by $0.05$ (QM40) and $0.21$ (QMugs) across seeds, while its SAIL fine-tuned versions vary by at most $0.01$.

    \begin{table}[h]
        \centering
        \caption{Mean ERIC over three training seeds for B3LYP/def2-SVP models trained on QM9.}
        \label{tab:seeds}
        \small
        \begin{tabular}{ll cc cc}
        \toprule
        \textbf{Model} & & \multicolumn{2}{c}{\textbf{QM40}} & \multicolumn{2}{c}{\textbf{QMugs}} \\
         & & base & SAIL & base & SAIL \\
        \midrule
        Coefficients   & $\mathbf{c}$ & $0.919 \pm 0.017$ & $0.756 \pm 0.001$ & $0.886 \pm 0.005$ & $0.771 \pm 0.009$ \\
        Density Matrix & $\mathbf{P}$ & $1.065 \pm 0.051$ & $0.696 \pm 0.010$ & $2.702 \pm 0.205$ & $0.745 \pm 0.004$ \\
        \bottomrule
        \end{tabular}
    \end{table}

    \paragraph{Convergence to the same solution.}
    A faster-converging initial guess is only useful if the SCF reaches the same solution. Table~\ref{tab:energies} compares the final energies of SAIL- and MINAO-initialized B3LYP calculations on QM40. Even the largest deviation, $1.3\times10^{-7}\:\textrm{Ha}$, lies four orders of magnitude below chemical accuracy ($1.6\times10^{-3}\:\textrm{Ha}$). The learned initialization therefore changes the path to the fixed point but not the fixed point itself.

    \begin{table}[h]
    \centering
    \caption{Deviation of the final energy of SAIL-initialized from MINAO-initialized SCF calculations on QM40 (B3LYP/def2-SVP).}
    \label{tab:energies}
    \small
    \begin{tabular}{ll cc}
    \toprule
    \textbf{Model} & & \textbf{MAE} [$\textrm{Ha}$] & \textbf{max.\ AE} [$\textrm{Ha}$] \\
    \midrule
    Coefficients   & $\mathbf{c}$ & $6.3\times10^{-11}$ & $5.5\times10^{-9}$ \\
    Density Matrix & $\mathbf{P}$ & $1.4\times10^{-10}$ & $1.3\times10^{-7}$ \\
    Fock           & $\mathbf{F}$ & $1.6\times10^{-10}$ & $1.3\times10^{-7}$ \\
    \bottomrule
    \end{tabular}
    \end{table}

    \paragraph{Per-molecule robustness.}
    Every SAIL-initialized calculation on QM40 and QMugs converges. For the density-matrix model, every molecule has an ERIC below one (maximum $0.93$). For the coefficient and Hamiltonian models, a single QM40 molecule with 16 heavy atoms exceeds one ($1.07$ and $1.21$, respectively); the density-matrix model reaches $0.93$ on the same molecule.

\end{document}